\documentclass{article}

\usepackage{microtype}
\usepackage{booktabs} 

\usepackage{hyperref}
\usepackage{bm}

\usepackage[accepted]{icml2024}

\usepackage{amsmath}
\usepackage{amssymb}
\usepackage{mathtools}
\usepackage{amsthm}

\usepackage{algorithm}
\usepackage[noend]{algpseudocode}

\algrenewcommand\algorithmicforall{\textbf{foreach}}
\algrenewcommand\algorithmicindent{.8em}

\usepackage[capitalize,noabbrev]{cleveref}

\theoremstyle{plain}

\theoremstyle{definition}

\theoremstyle{remark}

\usepackage[textsize=tiny]{todonotes}

\usepackage{url}
\usepackage{graphicx}
\usepackage{float}
\usepackage{subcaption}
\usepackage{dsfont}

\usepackage{float}
\usepackage{placeins}

\newcommand{\x}{\mathbf{x}}

\newcommand{\ba}{\mathbf{a}}

\newcommand{\I}{\mathbf{I}}

\newcommand{\W}{\mathbf{W}}

\icmltitlerunning{MONGOOSE: Path-wise Smooth Bayesian Optimisation via Meta-learning}

\begin{document}

\twocolumn[
\icmltitle{MONGOOSE: Path-wise Smooth Bayesian Optimisation via Meta-learning}




\begin{icmlauthorlist}
\icmlauthor{Adam X. Yang}{1}
\icmlauthor{Laurence Aitchison}{1}
\icmlauthor{Henry B. Moss}{2}
\end{icmlauthorlist}

\icmlaffiliation{1}{University of Bristol}
\icmlaffiliation{2}{University of Cambridge}

\icmlcorrespondingauthor{Adam X. Yang}{adam.yang@bristol.ac.uk}

\icmlkeywords{Machine Learning, ICML}

\vskip 0.3in
]



\printAffiliationsAndNotice{}  

\begin{abstract}
In Bayesian optimisation, we often seek to minimise the black-box objective functions that arise in real-world physical systems. 
A primary contributor to the cost of evaluating such black-box objective functions is often the effort required to prepare the system for measurement. 
We consider a common scenario where preparation costs grow as the distance between successive evaluations increases. 
In this setting, smooth optimisation trajectories are preferred and the jumpy paths produced by the standard myopic (i.e.\ one-step-optimal) Bayesian optimisation methods are sub-optimal. 
Our algorithm, MONGOOSE, uses a meta-learnt parametric policy to generate smooth optimisation trajectories, achieving performance gains over existing methods when optimising functions with large movement costs.
\end{abstract}

\section{Introduction}\label{sec:intro}


The task of optimising high-cost black-box functions is inescapable across science and industry. For many of these problems, evaluating the black-box is expensive, not due to the resources expended to take the measurement itself, but instead due to the substantial \textit{movement cost} required to transition the system to be ready for the next high-quality measurement --- a cost that increases with the distance (in the input space) between successive measurements. 
Example movement costs include: the financial outlay of moving mining machinery between drill sites when seeking areas dense in valuable ores \citep{jafrasteh2021objective}; the time taken for mixtures of chemicals to reach steady state  when trying to identify optimal mixtures \citep{teh2008droplet, rankovic2022bayesian}; or the effort required to reconfigure mechanical systems like particle accelerators \citep{roussel2021multiobjective} or heat exchangers \citep{paleyes2022penalisation}.

\begin{figure}%
\captionsetup[subfigure]{aboveskip=-3pt}
\subfloat[EI]{\includegraphics[height= 0.18\textwidth]{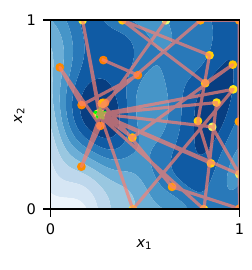}\label{fig:intro:ei}}%
\subfloat[EI per unit cost]{\includegraphics[height= 0.18\textwidth]{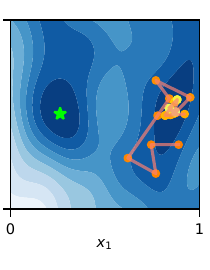}\label{fig:intro:eipu}}%
\subfloat[MONGOOSE]{\includegraphics[height= 0.18\textwidth]{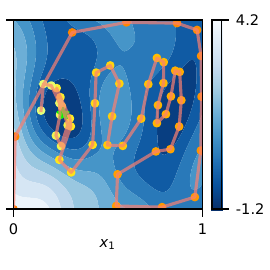}\label{fig:intro:mongoose}}%
\vspace{-0.3cm}
    \caption{ 50 minisation steps (orange dots to yellow dots) on a toy function (background). Standard BO with EI (a) incurs large movement costs, whereas EI per unit cost (b) fails to reach the global minima (star). Our non-myopic approach (c) finds the minima whilst following a smooth trajectory.
    }\label{fig:intro}
\end{figure}

Bayesian Optimisation \citep[BO]{shahriari2015taking} is a popular approach for black-box optimisation under constrained budgets. 
At first glance, BO appears to be a promising method for the problems above. 
However, standard BO is not designed for settings with movement cost constraints. 
As such, most methods, including those driven by acquisition functions such as Expected Improvement, favour reducing uncertainty in previously unexplored areas, a strategy that results in large jumps between successive evaluations. 
Therefore, while efficient in terms of the number of evaluations, standard BO is not efficient in terms of movement costs (see Figure \ref{fig:intro:ei}). 

At the same time, encouraging smooth optimisation paths by simply penalising large movements, e.g. considering the EI per unit movement cost (discussed in \cite{folch2022snake}), can lead to a failure to escape local optima (see Figure \ref{fig:intro:eipu}). 
This is due to the myopic nature of such an approach: it takes into account only the immediate benefit provided by making an evaluation. 
However, in order to acheive a global optimum by following a smooth evaluation path, we must accept the immediate sub-optimality of steadily traversing a low-quality region in order to access new promising areas, instead of jumping to the greedy solution --- a trade-off that will never be made under myopia. 

Successful movement-cost constrained BO thus requires non-myopic decision making.
Unfortunately, there has been limited success in developing non-myopic BO methods.
Solving the \textit{multi-step look-ahead problem} \citep{osborne2009gaussian}  is challenging since calculating non-myopic acquisition functions requires nested maximisations and expectations when conditioning the surrogate model over each future time step (see \cite{gonzalez2016glasses} for a discussion). 
Therefore the computational cost of existing non-myopic BO methods like \cite{jiang2020efficient} and \cite{lee2021nonmyopic} scales prohibitively for the longer time horizons ($\gg 10$) required for smooth global optimisation.

In this work, we propose a new algorithm, Meta-learning Of Non-myopic Global Optimisation fOr Smooth Exploration (MONGOOSE), for the optimisation of black-box functions under high movement costs (See Figure \ref{fig:intro:mongoose}). We sidestep the need to calculate non-myopic acquisition functions by leveraging recent developments in \textit{memory-based} optimisation to instead learn 
a non-myopic policy directly. In particular, we 
train a recurrent neural network to provide efficient cost-efficient optimisation over carefully crafted test functions based on samples from a Gaussian Process \citep[GP]{rasmussen2006gaussian}. 
Our chosen network architecture enjoys an inductive bias for smooth paths and our proposed loss function allows the degree of smoothness to be customised to the task at hand. 
Finally, we show that MONGOOSE improves over baselines for a variety of test functions.

\section{Background}

In this work, we seek to find the minimum of a smooth black-box function $f : \mathcal{X}\rightarrow \mathbb{R}$ over a compact search space $\mathcal{X}=[0,1]^d$ under a total evaluation budget of $T$ steps. Critically, we wish to perform this optimisation whilst incurring minimal cumulative moving cost $\mathcal{C}(\tau) = \sum_{t=0}^{T-1} \mathcal{C}(\x_{t},\x_{t+1})$. 
The cost function $\mathcal{C}:\mathcal{X}\times\mathcal{X}\rightarrow\mathbb{R}$ denotes the resources required to move between evaluations at $\textbf{x}_t$ and $\textbf{x}_{t+1}$. 
Our framework is agnostic to the exact form of the cost function, as long as it is differentiable, with the $L_1$ and $L_2$ distances being common examples.
The remainder of this Section details existing methods that are relevant for optimisation under movement costs, laying out important groundwork for our proposed MONGOOSE algorithm.

\subsection{Bayesian Optimisation}
In standard Bayesian Optimisation (BO) the goal is typically to  minimise $f$ in \emph{as few} evaluations as possible. Although this goal is not guaranteed to correspond to efficient optimisation under movement costs, we introduce it here as BO forms the basis for most existing methods for optimisation under movement costs.

BO achieves high data efficiency by using  previously collected function evaluations to build a probabilistic \textit{surrogate model} of the objective function. 
Typically GPs are used for these surrogates, however neural networks \citep{snoek2015scalable} and sparse GPs \citep{chang2022fantasizing,moss2023inducing} have also been considered.
This surrogate model is then used, through a search strategy known as an \textit{ acquisition function} $\alpha: \mathcal{X}\rightarrow \mathbb{R}$, to carefully select the next value of $\x$ at which to evaluate $f$, 
aiming to focus future resources promising areas of the space. Popular acquisition functions include those based on expected improvement \citep{jones1998efficient}, knowledge gradient \citep{frazier2008knowledge}, and Thompson sampling \citep{kandasamy2018parallelised}, as well a range of entropy-based methods \citep{hennig2012entropy, hernandez2014predictive, wang2017max, moss2021gibbon}.

\subsection{BO under movement costs}

A simple way to adapt BO to provide efficient optimisation with respect to movement costs is to incorporate these movement costs into its acquisition function. For instance the Expected Improvement per unit cost (EIpu) is defined as 
\begin{align}
\alpha_{\textrm{EIpu}}(\x_t, \x_{t+1}) = \alpha_{\textrm{EI}}(\x_{t+1})/(\gamma+\mathcal{C}(\x_{t},\x_{t+1})), \label{eq:eipu}
\end{align}
where $\alpha_{\textrm{EI}}$ the standard EI acqusition function and $\gamma$ is a small tuneable parameter (set as $\gamma=1$ by \cite{folch2022snake}). Unfortunately, EIpu heavily penalises the acquisition function away from the current location and often struggles to achieve global optimisation due to over-exploitation (recall Figure \ref{fig:intro}). 

The current state-of-the-art BO method for optimisation under movement costs is the Sequential Bayesian Optimisation via Adaptive Connecting Samples (SnAKe) of \cite{folch2022snake}. SnaKe follows the shortest path that connects a large number of promising regions, as identified through an approximate Thompson sampling scheme \citep{wilson2020efficiently,vakili2021scalable}. However, as demonstrated empirically by \cite{folch2022snake}, SnAKe has several shortcomings including the requirement of an additional heuristic to ensure that it avoids getting stuck in local modes and a drop in performance when considering higher dimensions and/or shorter time horizons.


\subsection{Memory-Based Optimisation}
There is a growing trend of training neural networks as black-box optimisers \citep{volpp2019meta, lange2022discovering, metz2022velo,chen2022scalable}; that is, teaching a network $M_{\theta}$ to take in $t$ previous evaluations and output a new promising location, i.e. $M_{\theta}:(\mathcal{X},\mathcal{Y})^t\rightarrow\mathcal{X}$ 
where $\theta$ denotes learnable weights. One immediate advantage over BO-based methods is that generating the next query point requires only a single forward pass of the network rather than the significant expense of fitting a GP and maximising an acquisition function. 
In particular, as the dimensionality of the problem increases, learning a decision policy directly side-steps the need to optimise an acquisition function in a high dimensional space.

\textbf{Network Architecture} 
A common choice for meta-optimisers is a memory-based network (e.g. recurrent neural networks) \citep{chen2021learning}, which typically stores an internal memory state that summarises the history of observations $\{(\x_{t'},f(\x_{t'})\}_{t'=1}^{t}$ and merges it with a current observation $(\x_t,f(\x_t)$ to produce a new location at which to evaluate $\x_{t+1}$. Such meta-trained meomory-based optimisers can memorise an effective adaptive search strategy based on the information learnt during meta-training, and reassuringly, they are known to achieve  close to (Bayes) optimal performance \citep{ortega2019meta, mikulik2020meta}.
Consequently, a widely used architecture for memory-based optimisers is the Long Short-Term Memory (LSTM) of \cite{hochreiter1997long} (see for example \cite{chen2017learning,mikulik2020meta,chen2021learning,ni2021recurrent} or \cite{metz2022velo}).

\textbf{Training Objective} To train a \textit{memory-based optimiser} it is common to use a meta-learning approach. More precisely, the network is trained to optimise a large set of objectives drawn from a distribution over functions which hopefully captures the true target objective,  e.g. \cite{chen2017learning} use functions sampled from a Gaussian process prior.

When measuring the performance of a particular optimiser over a fixed optimisation budget $T$, a natural non-myopic metric is to consider the overall improvement found by the optimiser. 
More precisely, we can write this training objective as
\begin{align} \label{eq:loss_orig}
    \mathcal{L}(\theta) = \mathbb{E}_f\left[f(\x_1) - \min_{t=1,...,T} f(\x_{t})\right],
\end{align}
where the expectation is taken with respect to a chosen prior over training functions $p(f)$. $\textbf{x}_t$ denotes the location of the $t^{th}$ evaluation chosen by our optimiser when applied to the function $f$, so $\textbf{x}_t$ is a function of both $\theta$ and the previous function evaluations, $f(\x_1),\dotsc,f(\x_{t-1})$.

Although this training objective (\ref{eq:loss_orig}) appears equivalent to the one discussed by \cite{chen2017learning}, 
due to a subtle implementation detail 
regarding the ``detaching'' of gradient terms related to non-myopia, the objective they actually optimise ends up being myopic. 
In contrast, we do not detach any gradients and instead use the full non-myopic objective for meta-training. Additional discussions and empirical results demonstrating a significant difference in performance between these two approaches are included in Appendix~\ref{app:myopic}.


\textbf{Meta-training} The objective in Equation~\ref{eq:loss_orig} is intractable, and therefore we use Monte-Carlo approximations during meta-training. At each optimisation step, we sample a set of $B$ functions $\{f_1,..,f_B\}$ from our prior distribution for $f$, and roll out our LSTM optimiser for each function, i.e we use the approximate objective
\begin{align}
    \mathcal{L}_{\textrm{MC}}(\theta) &= \frac{1}{B}\sum_{b=1}^B (f_b(\x_1) - \min_{t=1,...,T} f_b(\x_{t})). \label{eq:loss_mc}
\end{align}
During meta-training, we \emph{maximise} the objective with respect to the LSTM weights $\theta$ using a stochastic optimiser. 

If the memory-based optimiser is to be deployed on noisy objective functions, then we can simply add noise to the training functions to account for this, i.e. $f(\x) + \eta(\textbf{x})$, where $\eta$ is an arbitrary but known noise distribution. Note that standard BO methods are typically limited to Gaussian noise to ensure computational tractability. 

\section{MONGOOSE}

We now present our proposed algorithm, Meta-learning Of Non-myopic Global Optimisation fOr Smooth Exploration (MONGOOSE), which builds upon recent advances in memory-based meta learning and Bayesian optimisation to provide a black-box function minimiser that is efficient under large movement costs. At a high-level, MONGOOSE follows the ideas of \cite{chen2017learning} and meta-trains an LSTM, $M_\theta$, to optimise black-box functions. 
However, we introduce a number of key differences, including the use of a full non-myopic objective that incorporates moving cost, a better designed meta-training distribution, and a more efficient sampling and training scheme.

Our proposed MONGOOSE algorithm introduces three extensions to the work of \cite{chen2017learning} which improve the efficiency and applicability of memory-based optimisation. These are

\begin{enumerate}
    \item A training objective that encourages smooth optimisation paths.
    \item A new prior that generates more realistic training objective functions.
    \item A light-weight training scheme built upon efficient sampling methods.
\end{enumerate}

We expand on all three of these in the subsequent sections.



\subsection{Training Objective for Smooth Paths}
\label{subsec:loss}

We already have a non-myopic training objective for meta-training (\ref{eq:loss_orig}), however, it does not yet favour optimisation paths that incur minimal movement costs. 
Fortunately, we can easily incorporate a moving cost into our training objective as follows
\begin{align}
    \mathcal{L}_\text{div}(\theta) &= \frac{\mathcal{L}_{\textrm{MC}}(\theta)}{  1+ \alpha\sum_{1}^{T-1} c(\x_t, \x_{t+1})}. \label{eq:loss_div}
\end{align}
(We also consider an additive moving cost in Appendix~\ref{app:loss_add}.)
Here, $\alpha$ is a hyperparameter controlling the weight of moving cost, and $c(\cdot, \cdot)$ is a distance function.
Any differentiable function, $c(\x_t, \x_{t+1})$ is admissible.

\begin{figure*}[!t]
    \centering
    \includegraphics[width=\textwidth]{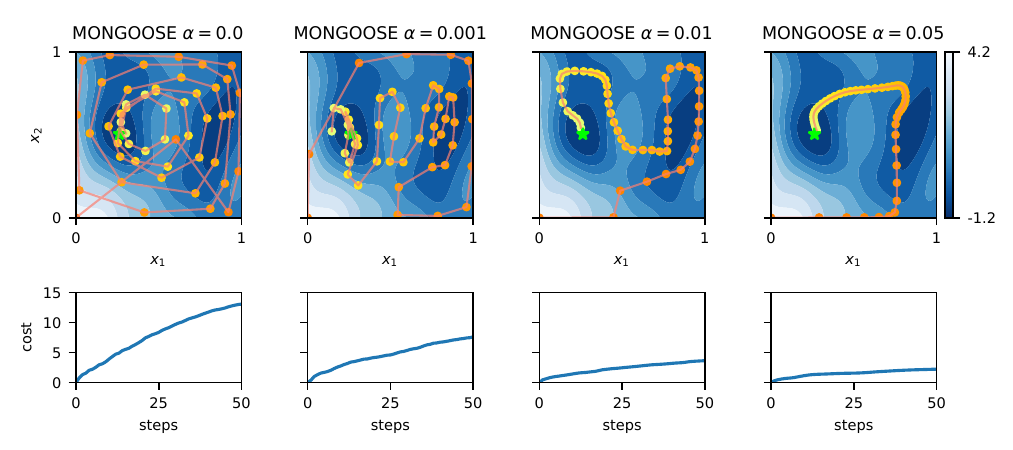}
    \caption{\textbf{Top}: trajectories MONGOOSE with different cost scalings on a single function sample from the meta-training distribution (background colour). Cost scalings $\alpha = 0.00, 0.01, 0.05$ from left to right as labelled on titles. Background with colour scale represent the function sample. Orange/yellow dots denote the evaluations chosen by each method, where darker colours (more orange) denote points earlier in the optimisation, and lighter colors (more yellow) denote points later in the optimisation. Consecutive evaluations are joined by lines. \textbf{Bottom}: $L_2$ distance (i.e. moving cost) to traverse each optimisation trajectory.}
    \label{fig:fourier_traj_div}
\end{figure*}

One of the key advantages of this training objective is that we can control the relative importance of moving costs using $\alpha$, an important degree of freedom that allows MONGOOSE to be customised to specific problem settings. In contrast, the current state-of-the-art SnAKe \cite{folch2022snake} lacks this flexibility. Figure~\ref{fig:fourier_traj_div} demonstrates that increasing $\alpha$ trades off cost for exploration, a trade that would be appropriate for problems where movement costs significantly dominate the cost of each function evaluation.

Interestingly, even without any moving penalty (i.e. setting $\alpha=0$) MONGOOSE still generates relatively smooth trajectories. We suspect this is an inductive-bias of memory-based models, where the memory-state 
may retain more information from the closest previous evaluation $\x_t$ 
(see Appendix~\ref{app:zero_cost_efficient} for a discussion).

\subsection{Injecting Global Structure}
\label{subsec:quad}

To guarantee performance at test time, it is of critical importance that the surrogate objective functions that we minimise at training-time are representative of the true test-time objective function. 
However, there is an emerging consensus that GP samples may not be representative of real-world objective functions.
First, \citet{le2021revisiting} and \citet{picheny2022bayesian} emphasise that real-world objectives often have a single global optimum, and ``global'' structure around that optimum. 
In contrast, functions sampled from GP priors with e.g.\ Mat\'ern or squared exponential kernels, have no global structure that extends beyond the GP lengthscale, and hence may have many comparably performing minima.  
Second, \cite{hvarfner2022pi} argue that global minima are likely to lie centrally in the search space (as the search space has been designed by experts to cover the likely value of the global optimum), while, due to the curse of dimensionality, GP samples have their minima focused along the edges of the search domains.

Therefore, to alleviate the shortcomings described above, we deviate from standard training function priors when training MONGOOSE. 
We sample a quadratic bowl and add this to the training functions sampled from GPs. This addition adds global structure to the training functions and increases the likelihood of having a single central global optima.
In particular, to generate a single training objective function, we first generate a sample $f$ from a GP prior, 
then add a randomly generated convex quadratic,
\begin{align} \label{eq:final_func}
    f_\text{quad}(\x) = f(\x) + \tfrac{1}{d} (\x-\ba)^T \W (\x-\ba) + c.
\end{align}
Here, $\textbf{W}$ is sampled from a Wishart distribution $\mathcal{W}(\tfrac{1}{d}\I,d)$ to ensure convexity, $\ba\sim U(0.2,0.8)^d$ 
to encourage a central minima, and $c = \tfrac{1}{8d} \sum_{i,j}[\W]_{ij}$ (half the expected maximum value of the quadratic) to ensure that the inclusion of the quadratic doesn't dramatically change the output range of the sampled functions. We found that including this global structure gives an improvement in optimisation performance downstream, especially in higher dimensions (see Appendix~\ref{app:glob_struct}).

\subsection{Meta-training By Fourier Features}
\label{subsec:RFF}

Recall that calculating our training objective (\ref{eq:loss_mc}) requires the evaluation of $K$ samples from a GP prior, each across $T$ locations. Previous meta-training approaches sample the GP exactly \citep{chen2017learning}, however, due to a Cholesky decomposition step \citep{diggle1998model}, this incurs a $O(T^3)$ cost which becomes prohibitively expensive for longer time-horizons.

A natural answer to these scalability issues is to rely instead on an approximate sampling schemes already commonly used throughout BO literature known as Random Fourier Features (RFF). In particular, it is well-known that for many common choices of kernels, GP samples can be expressed as a weighted sum of the kernel’s Fourier features \citep{rahimi2007random}. This sum can then be truncated to only its $M$ largest contributors, leaving approximate but analytically tractable samples that can be queried with only $O(MT)$ cost. See Appendix A of \cite{hernandez2014predictive} for full details. In our experiments, we found that using these approximate samples for training allowed a dramatic reduction in training costs without a loss in training stability or in the performance of the trained optimisers.




We can now summarise the full algorithm for MONGOOSE in Algorithm~\ref{alg:mongoose}. Note that the roll-out of MONGOOSE over the $B$ training functions (i.e. line 5 of Algorithm~\ref{alg:mongoose}) is entirely parallelisable.

\begin{algorithm}[ht]
\begin{algorithmic}[1]
\State Choose Horizon $H$, \# training steps $N$, Batch size $B$
\For {$n\in\{1,..,N\}$}  \textcolor{gray}{\Comment{Training loop}}
    \State Generate $B$ approximate GP samples $\{f_1,..,f_B\}$
    \State Add random quadratic effects $f_b \leftarrow f_b + f_{\textrm{quad}}$ 
    \For {$b\in\{1,..,B\}$} \textcolor{gray}{\Comment{Can be parallelised}}
        \For {$h\in\{1,..,H\}$}\hfill \textcolor{gray}{\Comment{Rollout}}
            \State$\textbf{x}_h=M_{\theta}(\{(\textbf{x}_i, f_b(\textbf{x}_i))\}_{i=1}^{h-1})$  
        \EndFor
    \EndFor
    \State Use the $B$ roll-outs to calculate $\mathcal{L}_{\textrm{div}}(\theta)$  \textcolor{gray}{\Comment{Eq. \ref{eq:loss_div}}}
    \State Backpropogate through $\mathcal{L}_{\textrm{div}}(\theta)$ and update $\theta$
\EndFor
\State \Return A trained MONGOOSE $M_{\theta}$
\end{algorithmic}
\caption{\label{alg:mongoose}Training MONGOOSE}
\end{algorithm}

\section{Related Work}

\textbf{Cost-constrained BO} There are many examples of BO where the cost of evaluations depends on their location (rather than the relative distance from previous evaluations we consider). In this popular setting, building a simple cost-weighted acquisition function like the EI per unit evaluation cost, can sometimes be an effective heuristic, e.g. when tuning the architecture of neural networks where certain design choices increase training times \citep{snoek2012practical} or when  multiple evaluation methods are available but each with differing costs, as arise in multi-task \citep{swersky2013multi}, multi-source \citep{poloczek2017multi} or multi-fidelity \citep{moss2021mumbo} optimisation. Unfortunately, as discussed above and demonstrated in our experiments, applying a simple cost-weighting idea (similar to that proposed by \cite{roussel2021multiobjective}) for the movement cost setting can lead to arbitrarily poor optimisation. Recently, \cite{lee2020efficient,lee2021nonmyopic} reformulated BO under location dependent costs as a constrained Markov decision process, trading a performance improvement over cost-weighted baselines for significant additional computational complexity. In other related work, \citet{ramesh2022movement} considers a similar movement penalty but in a specific contextual BO setting inspired by wind energy systems.

\textbf{Non-myopic BO} When performing global optimisation under a fixed evaluation budget, it should be advantageous to think non-myopically. Consequently, many non-myopic BO approaches have been proposed outside of the cost-constrained setting, ranging from cheap heuristics like GLASSES \citep{gonzalez2016glasses} and BINOCULARS \citep{jiang2020binoculars}, which approximate multi-step look-ahead as batch experimental design problem, to expensive approximations of optimal non-myopic policies \citep{jiang2020efficient,yue2020non,lee2021nonmyopic} suitable for shorter time-horizons (< 10 steps). Note that SnAKE can be interpreted as an extension of GLASSES  \citep{gonzalez2016glasses} to the movement constrained setting, achieving a degree of non-myopic decision making via constructing batches.

\textbf{Meta-learnt optimisers} \citet{chen2017learning}  meta-trained a long short-term memory (LSTM) network \citep{hochreiter1997long} over samples from a GP. 
However, their training framework involves conditional sampling of GPs which is both computational and memory intensive. 
Similarly, \cite{lange2022discovering} meta-learned an evolutionary strategy for BO rollout through an attention network, which itself is learned by another outer-loop evolutionary strategy.
\citet{volpp2019meta} amortised the acquisition function by a meta-learned neural acquisition function over GP a posterior, and subsequently learned a categorical policy on a grid of points through proximal policy optimisation \citep{schulman2017proximal}.
Meta-learning with memory-based agents also achieves state-of-the-art performance in many sequential decision making tasks \citep{ni2021recurrent}.
However, none of these existing methods support optimisation under large movement costs.

\section{Experiments}
\label{sec:experiments}

We now investigate the performance of MONGOOSE across three different settings: standard BO benchmark functions, across the extensive COCO testing suite\citep{finck2010real,hansen2021coco}, and on a real world example from \cite{folch2022snake}. For clarity, all our results follow a similar format, presenting regret against the movement costs incurred over $50$ (main text) and $100$ (Appendix \ref{app:100steps}) evaluations. All results are based on $50$ runs across different random seeds except for MONGOOSE which, due to computational considerations, was ran $10$ times for each experiment. 
Results on noiseless functions are included in the main text. See Appendix~\ref{app:noisy} for the corresponding results on noisy objective functions.

\subsection{Implementation Details}

\textbf{MONGOOSE} Our architecture comprises an LSTM with a hidden cell state dimension of 128, and a decoder with a sigmoid activation that maps hidden states to locations in the search space, with some additional design choices that were helpful in improving training stability. Firstly, we initialised MONGOOSE with a single evaluation at the origin $\x_0=\mathbf{0}$ (i.e a corner of the search space), with all subsequent evaluations chosen by the model. We found that starting with a randomly located evaluation could lead to less stable model fitting. Secondly, we also found curriculum learning \citep{bengio2009curriculum} to be important for stability, i.e. we began the optimisation process with shorter horizon lengths and gradually increased it to the desired longer horizons. Each curriculum phase comprised $5,000$ optimisation steps, with each training loss evaluation calculated using a new random batch of $128$ training functions. Back propagation through time is used to collect gradients \citep{werbos1990backpropagation}. Finally, we decayed the learning rate from $1\mathrm{e}{-3}$ to $1\mathrm{e}{-4}$ when the curriculum's horizon length reaches 40. 

To generate each function used to train MONGOOSE, we first sample a per-dimension lengthscale vector $\bm{\ell}\in\mathbb{R}^d$ from an inverse Gamma distribution with 99\% confidence interval at $[0.1,0.4]$, and then use this length-scale to build a GP with a M\`atern 5/2 kernel with unit variance from which we approximately sample using  $100$ RFFs. Our choice of randomly sampled lengthscale gives the GP sample variability across input dimensions while being realistic and covers a wide range of possible test functions. The source code for our experiments
has been made publicly available\footnote{\url{https://anonymous.4open.science/r/mongoose_submission-5131/}}.

\begin{figure*}[!t]
    \centering
    \includegraphics[width=\textwidth]{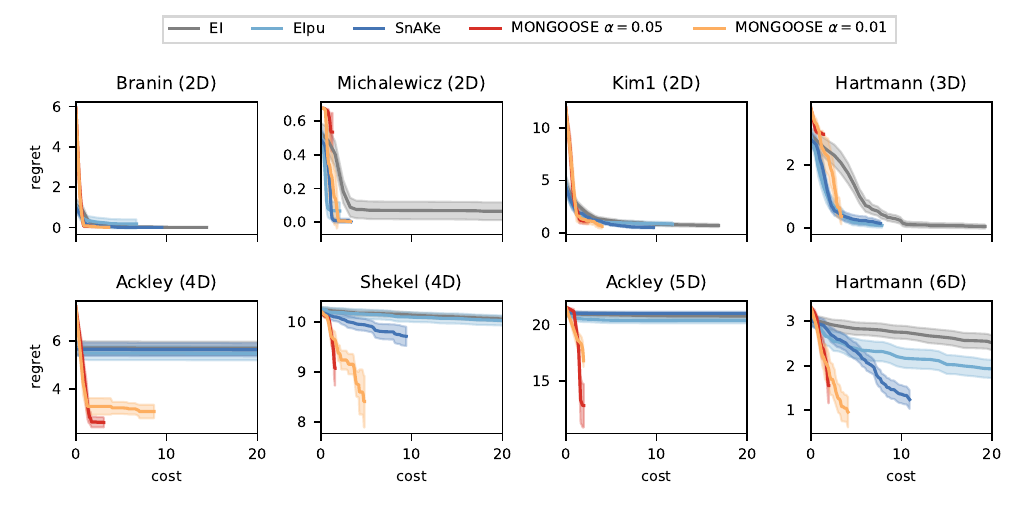}
    \caption{Regret versus cost on standard benchmark objective functions for two versions of MONGOOSE and BO baselines. We plot the mean and a $90\%$ confidence interval of regret for each method.}
    \label{fig:BO}
\end{figure*}

\textbf{Competitors} We use the implementation for EI, EIpu and SnAKe provided by \citet{folch2022snake}\footnote{\url{https://github.com/cog-imperial/SnAKe}} based on the BOTorch BO library \citep{balandat2020botorch}. We follow the recommendations of \cite{folch2022snake}, setting SnAKe's $\epsilon$-Point deletion scale to $\epsilon=0.1$ (a tune-able parameter that helps encourage global exploration) and EIpu's cost-scale coefficient to $\gamma=1$ (for other choices of $\gamma$, see Appendix~\ref{app:EIpu}).

Critical to the performance of BO methods, is access to an initial set of evaluations, from which reliable estimates of model parameters (e.g. lengthscales) can be calculated. Under movement costs, standard space-filling designs incur significant costs and so are likely sub-optimal, however, reliable estimates of model parameters are still required to ensure effective optimisation. We follow the setup of \cite{folch2022snake} and ``warm-start'' the BO methods (SnAKE, EI and EIpu) by providing them with a reasonable initialisation of GP model parameters (as calculated over an initial design of $10d$ points). As these evaluations are not used directly to fit surrogate models (only indirectly to provide an initial lengthscale), \cite{folch2022snake} chose not to include the cost of this design in the reported cost of their algorithm, a convention we also follow. In contrast, MONGOOSE starts from scratch from a single evaluation at the origin, i.e. with no warm-starting. Despite this substantial advantage given to the baseline methods, we will see that MONGOOSE still achieves superior performance.

\subsection{Bayesian optimisation benchmarks}

Firstly, we investigate the performance of MONGOOSE on standard BO benchmark functions 
as presented in Fig.~\ref{fig:BO}. 
In lower dimensions all algorithms perform similarly, however, when considering higher dimensions ($>3$), MONGOOSE consistently achieves lower regret with lower cost, a difference especially pronounced on the challenging highly multi-modal Ackley function. Note that these results match those claimed for EI, EIpu and SnAKe in Figure 11(b), Figure 12(b), Figure 13(b), Figure 14(b), and Figure 15(b) of \citet{folch2022snake}.

\begin{figure*}[!t]
    \centering
    \includegraphics[width=\textwidth]{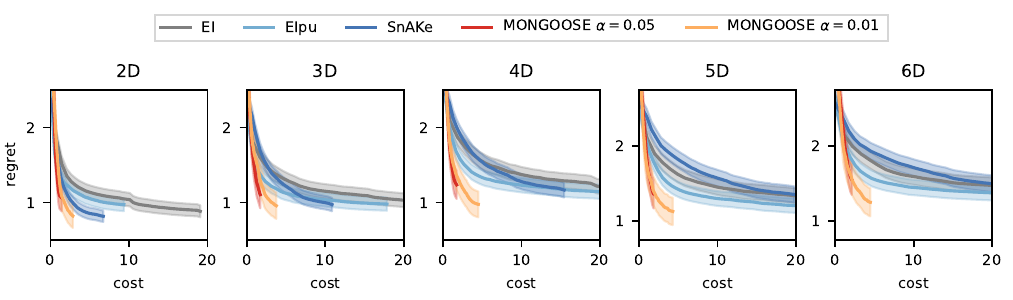}
    \caption{Regret against cost averaged across 24 Coco functions for a range of dimensions.}
    \label{fig:coco_avg}
\end{figure*}

\subsection{COCO test suite}

For a more thorough evaluation across different types of functions and across dimensions, we now consider the challenging COCO (COmparing Continuous Optimisers) test suite \citep{finck2010real,hansen2021coco}, a suite of 23 functions designed to benchmark black-box optimisers. Each function designed specifically to exhibit different attributes (e.g.\ multi-modality, low/high conditioning, weak/adequate global structure) and can be defined for arbitrary dimensions. 
We standardised these functions to make their values lie in a reasonable range (see Appendix~\ref{app:coco} for more details). 
The amortised results for dimensions two to six are included in Figure \ref{fig:coco_avg} (see Appendix~\ref{app:coco_50_orig}  for a per-function breakdown). 
MONGOOSE reliably achieves the best tradeoff between movement costs and regret across across all dimensions.
We believe that the poor performance of SnAKe in five and size dimensions is due to the requirement for a batch of points to achieve good coverage of the space, which becomes increasingly difficult in higher dimensions and under time horizons of only $50$. 
In Appendix \ref{app:100steps}, we show similar results for time horizons of $100$ evaluations.

\begin{figure*}[!t]
    \centering
    \includegraphics[width=\textwidth]{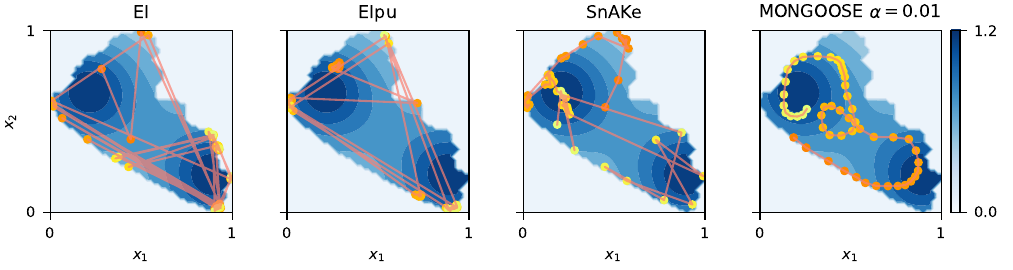}
    \caption{Optimisation trajectories generated when searching for contaminates across the Ypacarai Lake.}
    \label{fig:lake_traj}
\end{figure*}



The computational overhead incurred by MONGOOSE when optimising the COCO functions is around three orders of magnitude faster than achieved by the BO baseline methods  (see Table~\ref{tab:time}). Of course, MONGOOSE has the additional cost of requiring meta-training, however, as this takes less than 30 minutes on one RTX2080Ti (when considering a 50 step time horizon) and only needs to be performed once for each considered input dimensionality (i.e. not for each objective function), we do not consider meta-training a serious computational bottleneck.

\setlength{\tabcolsep}{4pt}
\begin{table}[!t]
    \centering
    \caption{Averaged time for 50 optimisation steps (in seconds) over the COCO test suite.}
    \begin{tabular}{cccccc}
        \toprule
        methods & 2D & 3D & 4D & 5D & 6D\\
        \midrule
        EI & 33 & 33 & 35 & 36 & 36 \\
        EIpu & 33 & 34 & 36 & 36 & 37 \\
        SnAKe & 23 & 33 & 46 & 70 & 92 \\
        MONGOOSE & \textbf{0.02} & \textbf{0.02} & \textbf{0.02} & \textbf{0.02} & \textbf{0.02}  \\
        \bottomrule
    \end{tabular}%
    \label{tab:time}
\end{table}

\subsection{Real world example}

\begin{figure}[!t]
    \centering
    \includegraphics[width=.5\textwidth]{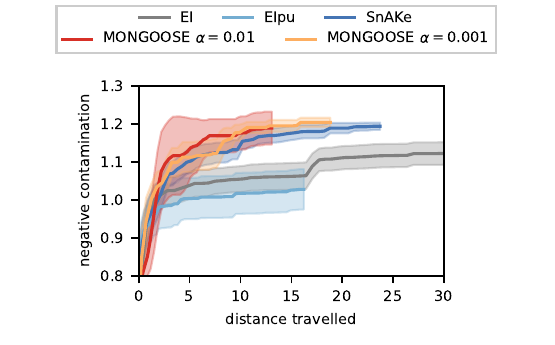}
    \caption{The maximum contamination found against distance travelled by EI, EIpu, SnAKe and MONGOOSE.}
    \label{fig:lake_regret}
\end{figure}

For our final example, we turn to the Ypacarai Lake problem \citep{samaniego2021bayesian,folch2022snake} --- a real world black-box optimisation problem that suffers from substantial movement costs. Here, the task is to direct an autonomous surface vehicle to locate contamination sources in the lake, thus travelling a minimal distance is preferred to minimise time and energy consumption. The ground-truth contamination levels over the lake are given over a fine grid. For the BO baselines of EI, EIpu and SnAKe, we use this pre-specified grid as their search space, whereas for MONGOOSE, we project the locations to closest grid point and evaluate the objective at the projected location. Figure~\ref{fig:lake_traj} compares the trajectories from a single run of EI, EIpu, SnAKe and MONGOOSE, demonstrating that MONGOOSE with $\alpha=0.01$ is able to generate an entirely smooth trajectory that explores both modes. Figure~\ref{fig:lake_regret} shows the maximum contamination found  against distance travelled by different methods. 

\section{Discussion}
In this work, we developed a  memory-based meta-learning approach for the optimisation of black box functions where inputs incur large costs. 
and our results showed MONGOOSE performs better than competing methods (EI, EIpu and SnAKe) over horizons of 50-100 steps, especially in higher dimensions.

In future work we will investigate the use of dimensional agnostic architectures to avoid the need to train separate network for objective functions with different input dimensions.  Attention-based architectures \citep{lee2019set, simpson2021kernel} may provide a solution, with the additional
benefit of being invariant to the ordering of query points (a property that should hold for Bayes optimal agents, see \cite{ortega2019meta} or \cite{mikulik2020meta}).
We will explore the use of more flexible functions approximators to construct  meta-training distributions \citep{aitchison2021deep,ober2021variational,yang2021theory}.  
Another open question is how to extend memory-based optimisers to non-Euclidean search spaces, a jump recently made by BO in the context of gene design \cite{moss2020boss},
molecular search \cite{ griffiths2022data,griffiths2022gauche}
and combinatorial optimisation \cite{deshwal2021mercer}.

\nocite{langley00}

\bibliography{ref}
\bibliographystyle{icml2024}

\newpage
\appendix
\onecolumn

\section{Alternative objectives}

\subsection{Additive moving cost}
\label{app:loss_add}
\begin{figure}[H]
    \centering
    \includegraphics{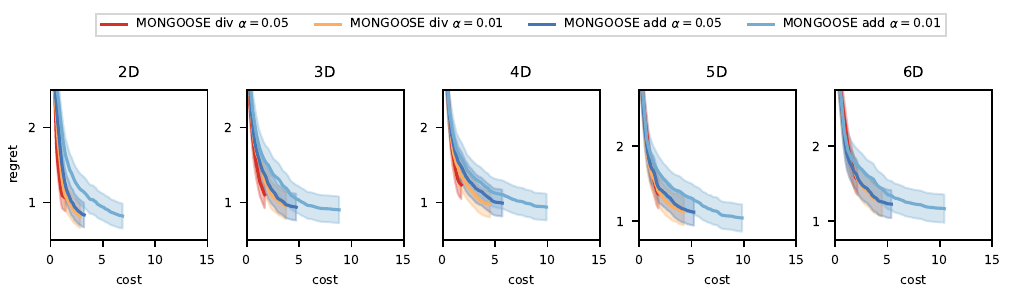}
    \caption{Comparison of $\mathcal{L}_\text{div}(\theta)$ versus $\mathcal{L}_\text{add}(\theta)$ averaged across the COCO benchmark. Individual plots for each COCO function are shown in Figure~\ref{fig:coco_50_1_divadd},\ref{fig:coco_50_2_divadd},\ref{fig:coco_50_3_divadd},\ref{fig:coco_50_4_divadd}.}
    \label{fig:divadd_coco}
\end{figure}

In the main text, we considered incorporating moving cost through division,
\begin{align}
    \mathcal{L}_\text{div}(\theta) &= \frac{\mathcal{L}(\theta)}{  1+ \alpha\sum_{t} c(\x_t, \x_{t+1})}.
\end{align}
Here, we consider the alternative option to add the cost
\begin{align}
    \mathcal{L}_\text{add}(\theta) &= \mathcal{L}(\theta) + \alpha \sum_{t=1}^{H-1} c(\x_t, \x_{t+1}),\label{eq:loss_add},
\end{align}
however, this is not the ideal choice for black-box functions since the choice of cost scaling $\alpha$ in the additive case needs to be proportional to the scaling of the function and so is difficult to predetermine.
Figure~\ref{fig:divadd_coco} compares their performance on normalised COCO functions, notice that they perform similarly under slightly different choices of cost scaling $\alpha$. In particular, $\mathcal{L}_\text{div}$ with $\alpha=0.001$ gives very similar performance as $\mathcal{L}_\text{add}$ with $\alpha=0.01$, and $\mathcal{L}_\text{div}$ with $\alpha=0.01$ is similar to $\mathcal{L}_\text{add}$ with $\alpha=0.05$.


\subsection{The myopic objective} \label{app:myopic}
The objective we use in our meta-training, as defined in Equation~\ref{eq:loss_orig},  is given by
\begin{align*}
    \mathcal{L}(\theta) = \mathbb{E}_f\left[f(\x_1) - \min_{t=1,...,T} f(\x_{t})\right].
\end{align*}
This is the expected improvement over our prior with respect to the minimum function value reached during a trajectory of $T$ steps.
It is worth noting that this loss can be expressed as a cumulative sum of improvement, which has the same form as the `observed improvement' proposed by \citet{chen2017learning},
\begin{align} 
    \mathcal{L}(\theta) &= \mathbb{E}_f\left[ \sum_{t=1}^T \max(\min_{t'=1,...,t-1} f(\x_{t'}) - f(\x_{t}),0) \right]. \nonumber
\end{align}
However, when optimising this loss, \citet{chen2017learning} detached the previous best function value $\min_{t'=1,...,t} f_k(\x_{t'})$ during back-propagation, effectively making the objective myopic. In contrast, we do not detach gradients and calculate the loss exactly as it is written. 
To aid intuition, consider a horizon length of 2 with a single training function and $\x_0=\mathbf{0}$. Under these assumption, our objective becomes 
\begin{align} 
    \mathcal{L}_\text{OI}(\theta) &=  \max(f(\x_0) - f(\x_{1}),0) + \max(\underbrace{\min\{f(\x_{0}),f(\x_1)\}}_{\text{detach}} - f(\x_2),0). \nonumber
\end{align}
Now consider detaching the gradient of $f(\x_1)$ from the second term. This leads to myopia because, when updating $\x_1$, its only contribution now comes from the first term which is only a one-step (i.e. myopic) improvement $\max(f(\x_0) - f(\x_{1}),0)$. In contrast, a truly non-myopic approach should consider the effect of changing $\x_1$ on all subsequent improvements.

\begin{figure}[H]
    \centering
    \includegraphics{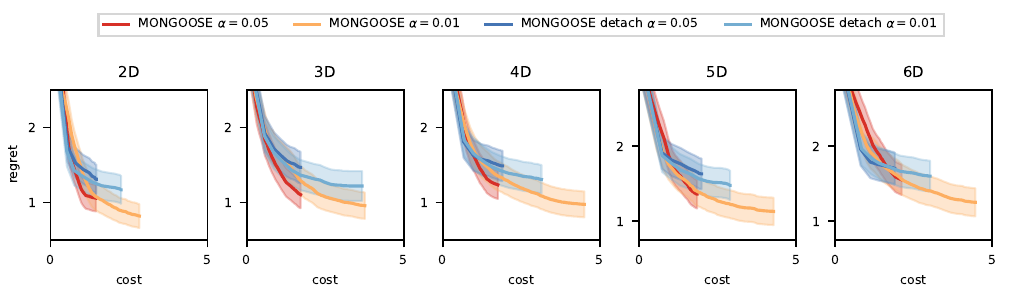}
    \caption{Comparing the non-myopic (red/orange) versus myopic (blues) meta-training objectives averaged across the COCO benchmark, with $\alpha=0.01, 0.05$. Individual plots for each COCO function are shown in Figure~\ref{fig:coco_detach_1},\ref{fig:coco_detach_2},\ref{fig:coco_detach_3},\ref{fig:coco_detach_4}.}
    \label{fig:coco_detach}
\end{figure}

From a more practical perspective, we saw a significant performance degradation when mimicking the gradient detaching of \cite{chen2017learning}, as shown in Figure~\ref{fig:coco_detach}.

\FloatBarrier
\newpage
\section{Effect of injecting global structure}
\label{app:glob_struct}
\begin{figure}[H]
    \centering
    \includegraphics{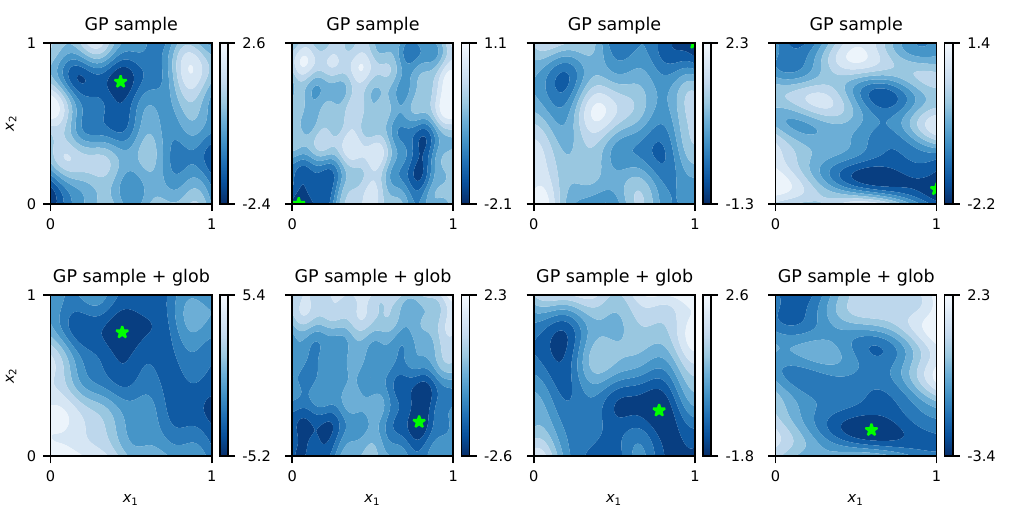}
    \caption{\textbf{Top}: original GP samples obtained from a M\`atern 5/2 kernel. \textbf{Bottom}: The same samples injected with randomly sampled global structure.}
    \label{fig:fourier_globstruct}
\end{figure}

\begin{figure}[H]
    \centering
    \includegraphics{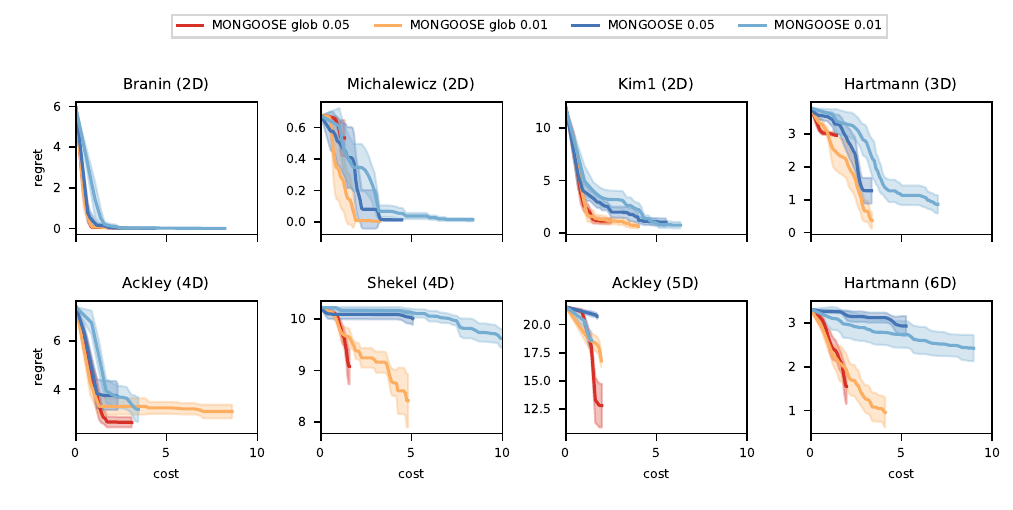}
    \caption{Investigating the effect of adding global structure during meta-training on standard BO benchmarks.}
    \label{fig:globstruct_bench}
\end{figure}

We now illustrate the effect  of injecting the global structure we described in the main text into GP samples as Figure~\ref{fig:fourier_globstruct}. 
The major effects of adding global structure include (1) moving global optimum from corners and edges towards the centre and (2) eliminating some modes at corners. Adding this global structure boosts the performance on standard BO benchmarks especially in 4D-6D functions (see Figure~\ref{fig:globstruct_bench}). When averaged across the COCO benchmark, MONGOOSE trained without global structure consistently achieves lower regret with a lower moving cost, and the advantage grows as dimensionality increases.

\begin{figure}[H]
    \centering
    \includegraphics{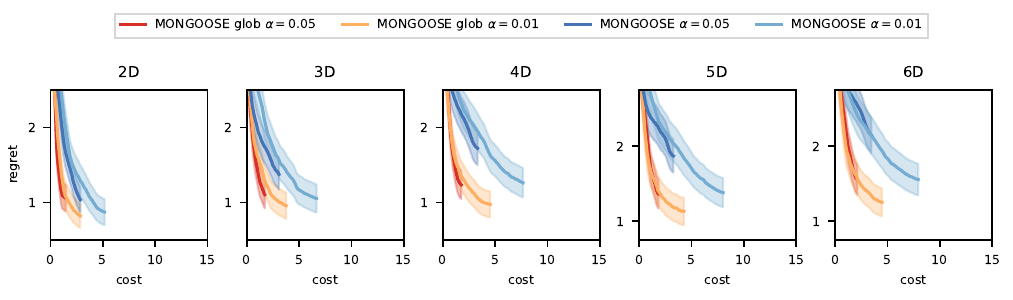}
    \caption{Investigating the effect of adding global structure (red/orange) against  standard GP sample (blue) during meta-training averaged across the COCO benchmark, with $\alpha=0.01, 0.05$. Individual plots for each COCO function are shown in Figure~\ref{fig:coco_50_1_glob},\ref{fig:coco_50_2_glob},\ref{fig:coco_50_3_glob},\ref{fig:coco_50_4_glob}.}
    \label{fig:globstruct_coco_avg}
\end{figure}

\FloatBarrier
\newpage
\section{Inductive bias of memory-based meta-optimisers}
\label{app:zero_cost_efficient}
\begin{figure}[H]
\captionsetup[subfigure]{aboveskip=-3pt}
\centering
\subfloat[\textbf{Top}: trajectories of EI, EIpu, SnAKe, colour scale same as in Figure~\ref{fig:intro}. \textbf{Bottom}: cumulative $L_2$ cost along the trajectory.]{\includegraphics[width=\textwidth]{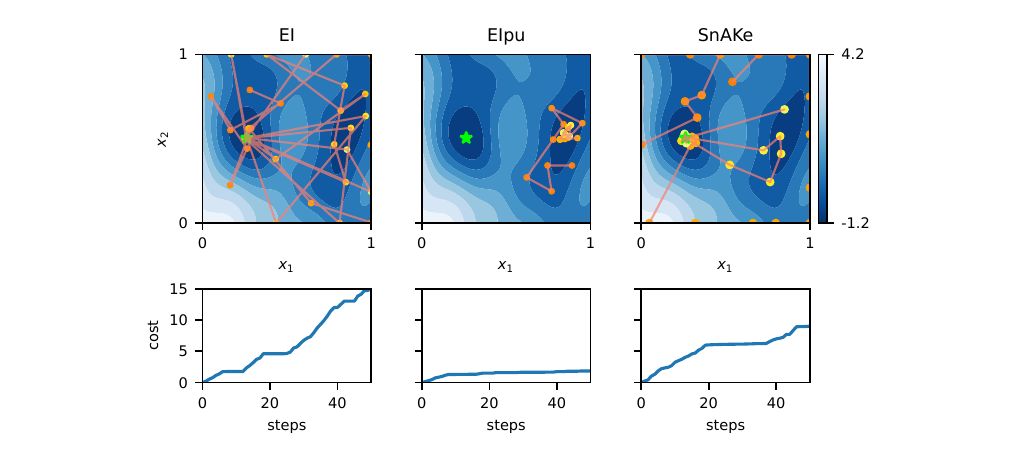}\label{fig:fourier_traj_app}}%

\subfloat[\textbf{Top}: trajectories of MONGOOSE with $\alpha=0, 0.001, 0.01, 0.05$, colour scale same as in Figure~\ref{fig:fourier_traj_div}. \textbf{Bottom}: cumulative $L_2$ cost along the optimisation trajectory.]{\includegraphics[width=\textwidth]{figures/fourier_div_217.pdf}\label{fig:fourier_traj_all}}%
\vspace{-0.3cm}
    \caption{ Comparing trajectories and costs from a single run of EI, EIpu, SnAKe, and MONGOOSE with $\alpha=0, 0.001, 0.01, 0.05$.
    }\label{fig:fourier_traj_div_app}
\end{figure}

We found that meta-trained memory-based optimisers using the non-myopic objective (Eq.~\ref{eq:loss_orig} and Eq.~\ref{eq:loss_div}) have an inductive bias of generating smooth trajectories with low cost. As illustrated in Figure~\ref{fig:fourier_traj_div_app}, even with $\alpha=0$, the trajectory of MONGOOSE is than the jumpy trajectory of EI (Figure~\ref{fig:fourier_traj_app}). Further evidence is provided in Figure~\ref{fig:zero_coco}, which shows MONGOOSE $\alpha=0$ outperforming EI, EIpu and SnAKe in terms of averaged regret versus cost on the COCO benchmark (for dimensions higher than 2D). 
We suspect that MONGOOSE's inducitve bias for smooth paths is due to hidden states in memory-based learners containing more information from closest previous steps, thus biasing the output to lie close to previous outputs.

\begin{figure}[H]
    \centering
    \includegraphics{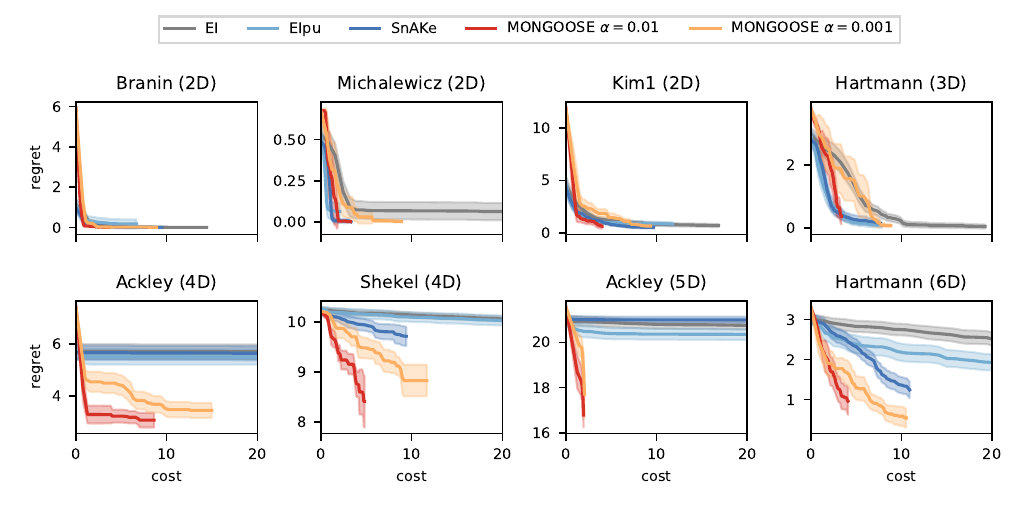}
    \caption{Comparison of EI, EIpu, SnAKe, MONGOOSE with $\alpha=0.001$ and $\alpha=0$ on standard BO benchmarks.}
    \label{fig:bench_50_zero}
\end{figure}

\begin{figure}[H]
    \centering
    \includegraphics{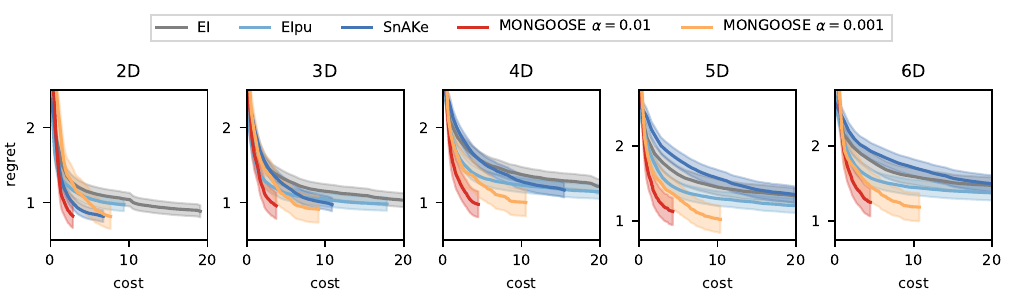}
    \caption{Comparison of EI, EIpu, SnAKe, MONGOOSE with $\alpha=0.01$ and $\alpha=0.001$ averaged across the COCO benchmark. Individual plots for each COCO function are shown in Figure~\ref{fig:coco_50_1_zero},\ref{fig:coco_50_2_zero},\ref{fig:coco_50_3_zero},\ref{fig:coco_50_4_zero}.}
    \label{fig:zero_coco}
\end{figure}

\FloatBarrier
\newpage
\section{Experiments on noisy functions}
\label{app:noisy}
\begin{figure}[H]
    \centering
    \includegraphics{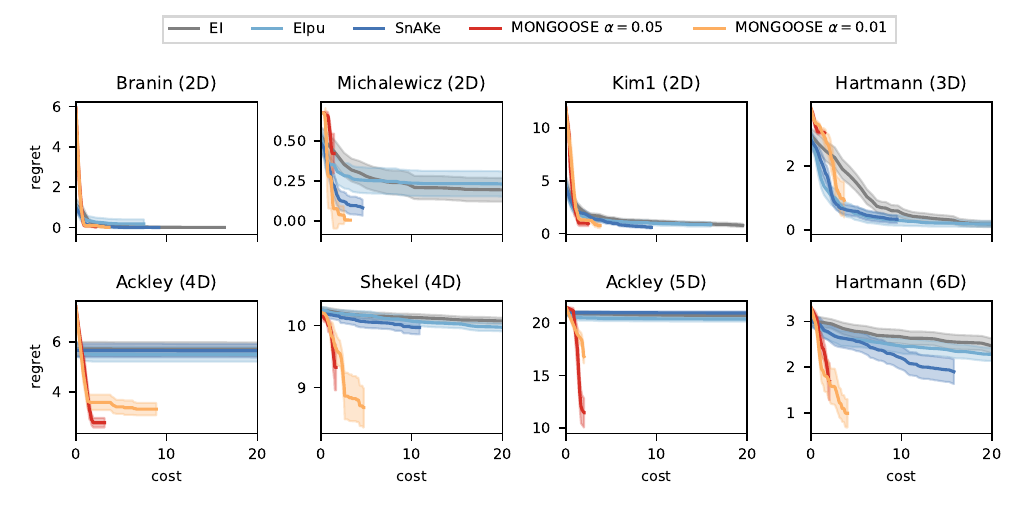}
    \caption{Comparison of EI, EIpu, SnAKe, MONGOOSE with $\alpha=0.01$ and $\alpha=0.05$ on standard BO benchmarks with observation noise $\eta\sim\mathcal{N}(0,0.1)$.}
    \label{fig:bench_50_noisy_0.1}
\end{figure}

\begin{figure}[H]
    \centering
    \includegraphics{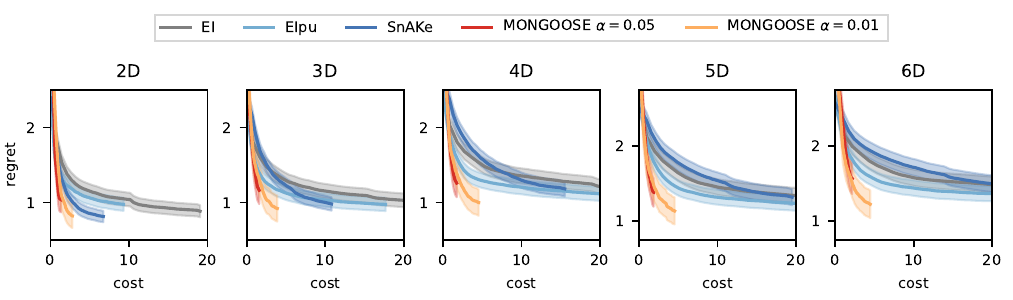}
    \caption{Comparison of EI, EIpu, SnAKe, MONGOOSE with $\alpha=0.01$ and $\alpha=0.05$ averaged across the COCO benchmarks with observation noise $\eta\sim\mathcal{N}(0,0.1)$. Individual plots for each COCO function are shown in Figure~\ref{fig:coco_50_1_noisy_0.1},\ref{fig:coco_50_2_noisy_0.1},\ref{fig:coco_50_3_noisy_0.1},\ref{fig:coco_50_4_noisy_0.1}.}
    \label{fig:coco_50_noisy_0.1}
\end{figure}

In the main text, we presented results on noiseless functions. Here, we consider adding Gaussian observation noise to function evaluations. Specifically, we sample noise $\eta\sim\mathcal{N}(0,\sigma^2)$, and let the model (GP for EI, EIpu and SnAKe; LSTM for MONGOOSE) observe new evaluation pair $(\x_t,f(\x_t)+\eta)$, and choose next evaluation location based on noisy observations. 
when computing the final regret for all methods at test time, we still use the true function value without observation noise.
The results for $\sigma^2=0.1$ on standard BO benchmarks are shown in Figure~\ref{fig:bench_50_noisy_0.1} and on the COCO benchmark are shown in Figure~\ref{fig:coco_50_noisy_0.1}

\FloatBarrier
\newpage
\section{Experiments for 100 steps horizon}
\label{app:100steps}
\begin{figure}[H]
    \centering
    \includegraphics{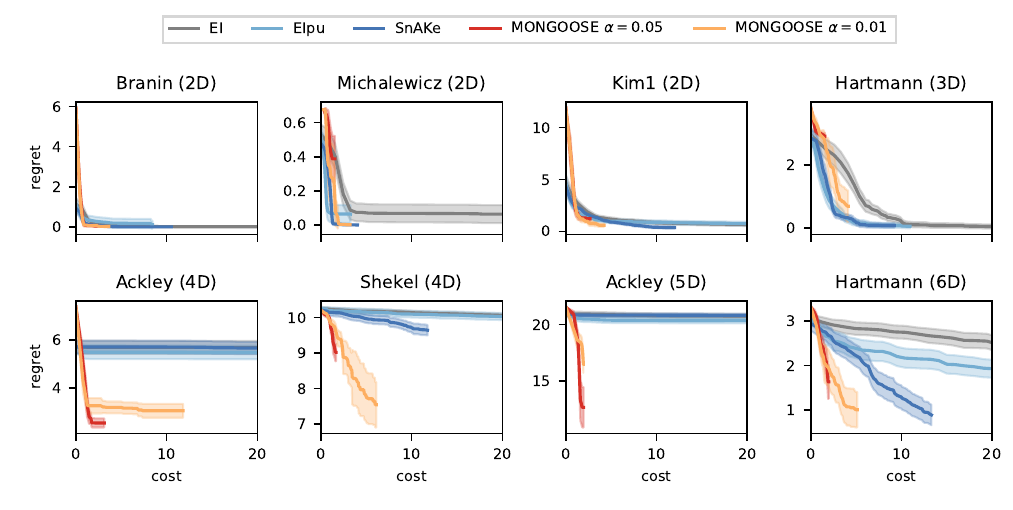}
    \caption{Comparison of EI, EIpu, SnAKe, MONGOOSE with $\alpha=0.01$ and $\alpha=0.05$ on standard BO benchmarks for a horizon of 100 steps.}
    \label{fig:bench_100}
\end{figure}

\begin{figure}[H]
    \centering
    \includegraphics{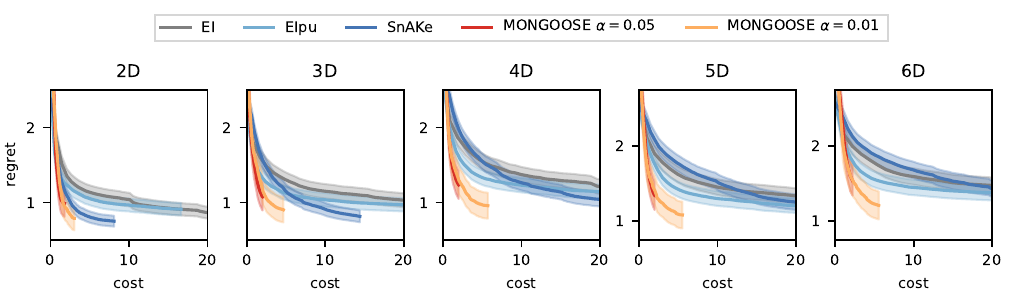}
    \caption{Comparison of EI, EIpu, SnAKe, MONGOOSE with $\alpha=0.01$ and $\alpha=0.05$ averaged across the COCO benchmarks for a horizon of 100 steps. Individual plots for each COCO function are shown in Figure~\ref{fig:coco_100_1},\ref{fig:coco_100_2},\ref{fig:coco_100_3},\ref{fig:coco_100_4}.}
    \label{fig:coco_100}
\end{figure}

In the main text, we showed results for a horizon of 50 steps. Here, we show the results for a horizon of 100 steps on standard benchmarks in Figure~\ref{fig:bench_100} and on COCO benchmarks Figure~\ref{fig:coco_100}. Our conclusions from the main paper still hold, although SnAKe does perform noticeable better on Hartmann 6D as well as all COCO functions, which is its expected behaviour as the number of steps in a BO loop grows \citep{folch2022snake}.

\FloatBarrier
\newpage
\section{EI per unit cost}
\label{app:EIpu}
\begin{figure}[H]
    \centering
    \includegraphics{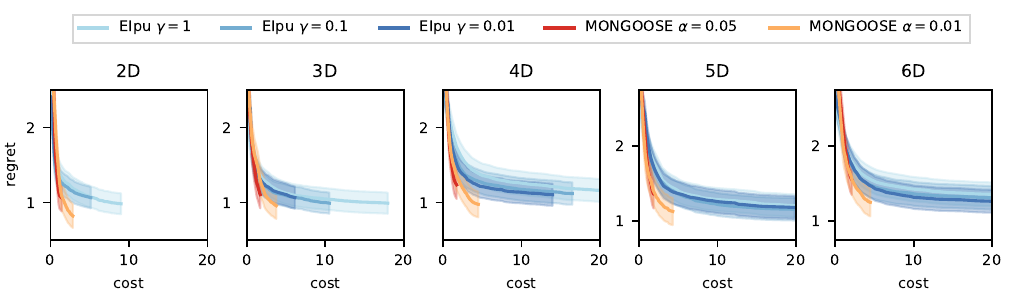}
    \caption{Comparison of EIpu with $\gamma=0.01, 0.1, 1$, and MONGOOSE with $\alpha=0.01,0.05$, averaged across the COCO benchmarks for a horizon of 100 steps. Individual plots for each COCO function are shown in Figure~\ref{fig:coco_eipu_1},\ref{fig:coco_eipu_2},\ref{fig:coco_eipu_3},\ref{fig:coco_eipu_4}.}
    \label{fig:coco_eipu}
\end{figure}

In this section we investigate the effect of the hyperparameter $\gamma$ in EI per unit cost (EIpu). Recall EIpu is defined as
\begin{align*}
    \text{EIpu}(x)=\frac{\text{EI}(\x)}{\gamma+1}.
\end{align*}
In the main text, we chose $\gamma=1$ following \citep{folch2022snake}.  As demonstrated in Figure~\ref{fig:coco_eipu}, EIpu with $\gamma=1,0.1,0.01$ are all outperfomed by MONGOOSE

\section{COCO functions}
\label{app:coco}
\begin{figure}[H]
    \centering
    \includegraphics{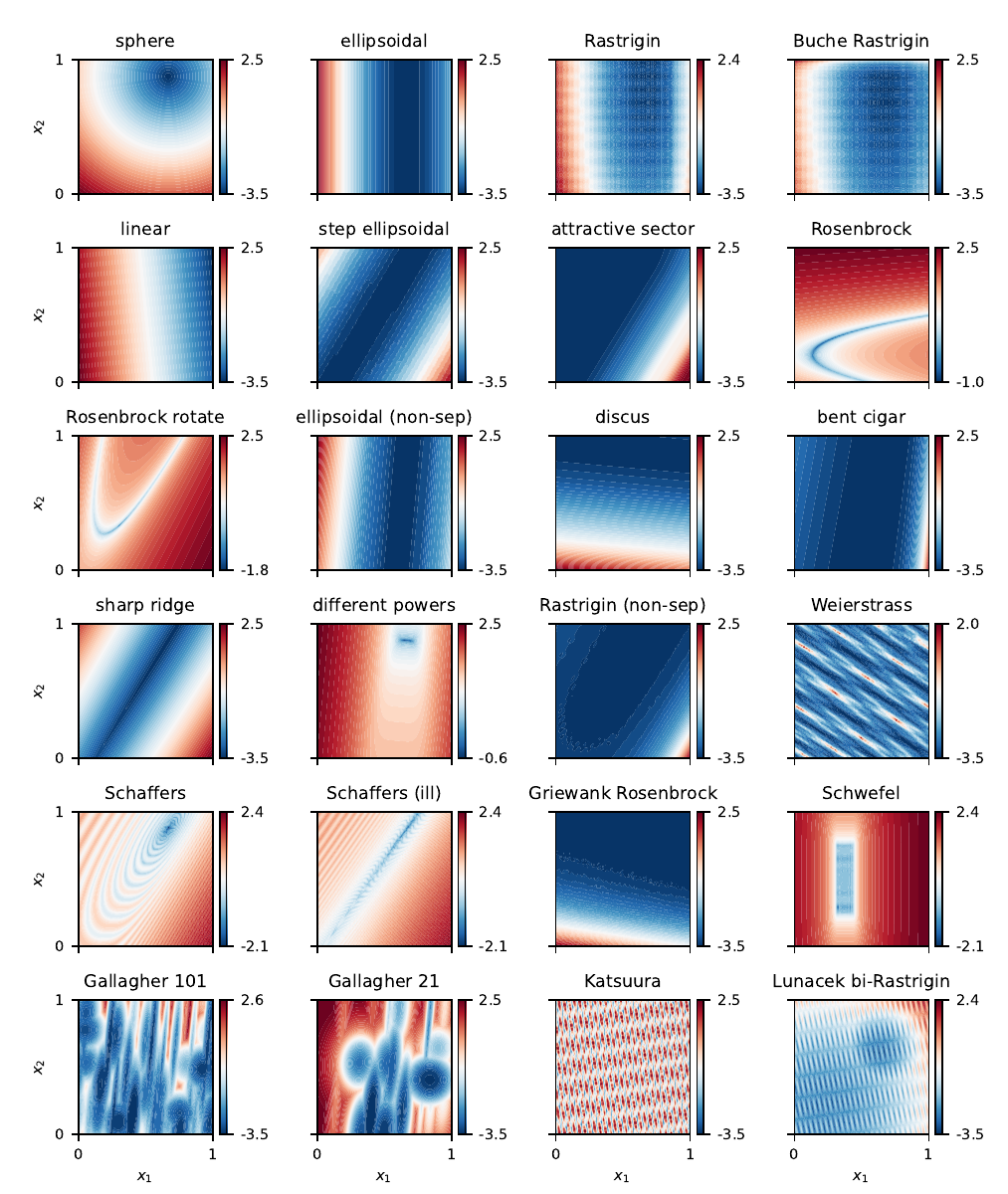}
    \caption{Plots of all 24 functions (noramlised) in the COCO benchmark.}
    \label{fig:coco_all}
\end{figure}

There are a total of 24 functions in the COCO benchmark \citet{finck2010real,hansen2021coco}, all of them are positive and have a known global minima with a corresponding minimum function value. Many have random parameters that we can sample to generate slightly different but similar functions. Since not all functions have this randomness and the random parameters are usually just rotations in the input space, we fixed all random parameters for our tests. 
One potential issue with functions in this benchmark is their outputs have vastly different ranges, for example, the ellipsoidal function (2D) ranges from 0 to $3\mathrm{e}{7}$ (\citet[p.~10]{finck2010real}), the Rastrigin function (2D) ranges from 0 to 800 (\citet[p.~15]{finck2010real}), the (log) Rosenbrock function (2D) ranges from 0 to 4 (\citet[p.~40]{finck2010real}), etc. Therefore, we chose to standardise these functions 
\begin{align*}
    \tilde{f}(\x) = \frac{f(\x)}{\max_\x f(\x) } \times 6 - 3 + f_\text{opt},
\end{align*}
where $\max_\x f(\x)$ is obtained through random search. Following, \citep{finck2010real} we add $f_\text{opt} \sim U[0,1]$ for additional randomness of the optimum value. Figure~\ref{fig:coco_all} shows plots for all functions after normalisation in the COCO benchmark. Note that since we are plotting with a grid of points, they might not cover the exact minimas/maximas especially when they are in a thin valley, so the minimums/maximums on the colourscales do not represent the exact minimum/maximum values of functions.
As described in the main text, for all experiments on the COCO benchmark, we meta-train 10 different MONGOOSE with 10 seeds, and we run EI, EIpu and SnAKe with 50 seeds each.

\FloatBarrier
\newpage
\section{COCO individual regret plots}
\label{app:coco_50_orig}

In this final section, we present individual regret plots for each of the 24 COCO functions \citep{finck2010real,hansen2021coco}, which are split into four plots of 6 functions for each setting above with an averaged COCO benchmark plot. We set the same y-axis scale across all regret plot to more easily see the results on which functions contribute more to the averaged differences in regret versus cost.



\begin{figure}
    \centering
    \includegraphics{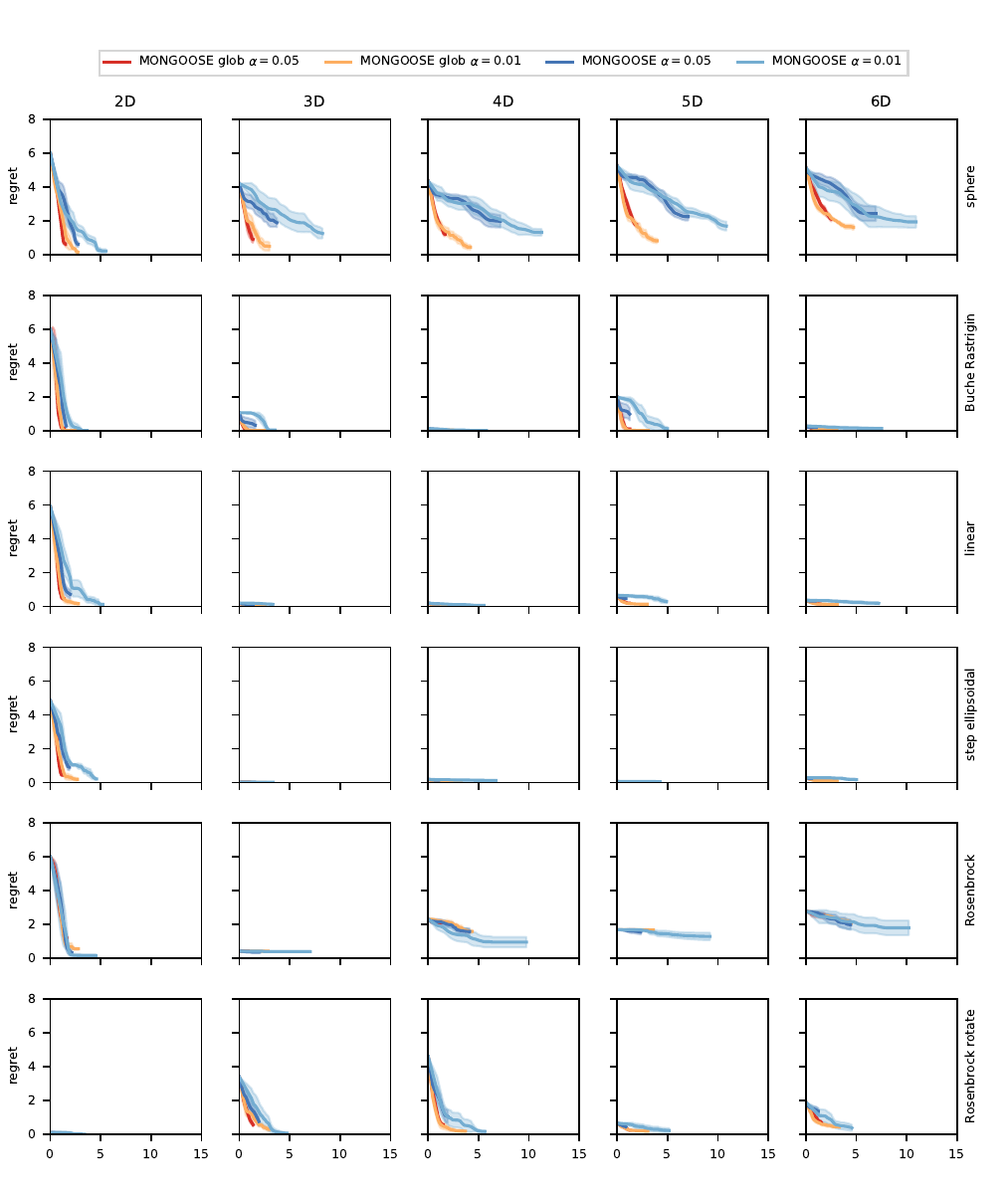}
    \caption{Individual COCO plots for Figure~\ref{fig:globstruct_coco_avg}. COCO functions 1-6: sphere function, ellipsoidal function, Rastrigin function, B\"{u}che-Rastrigin function, linear slope, step ellipsoidal function.}
    \label{fig:coco_50_1_glob}
\end{figure}

\begin{figure}
    \centering
    \includegraphics{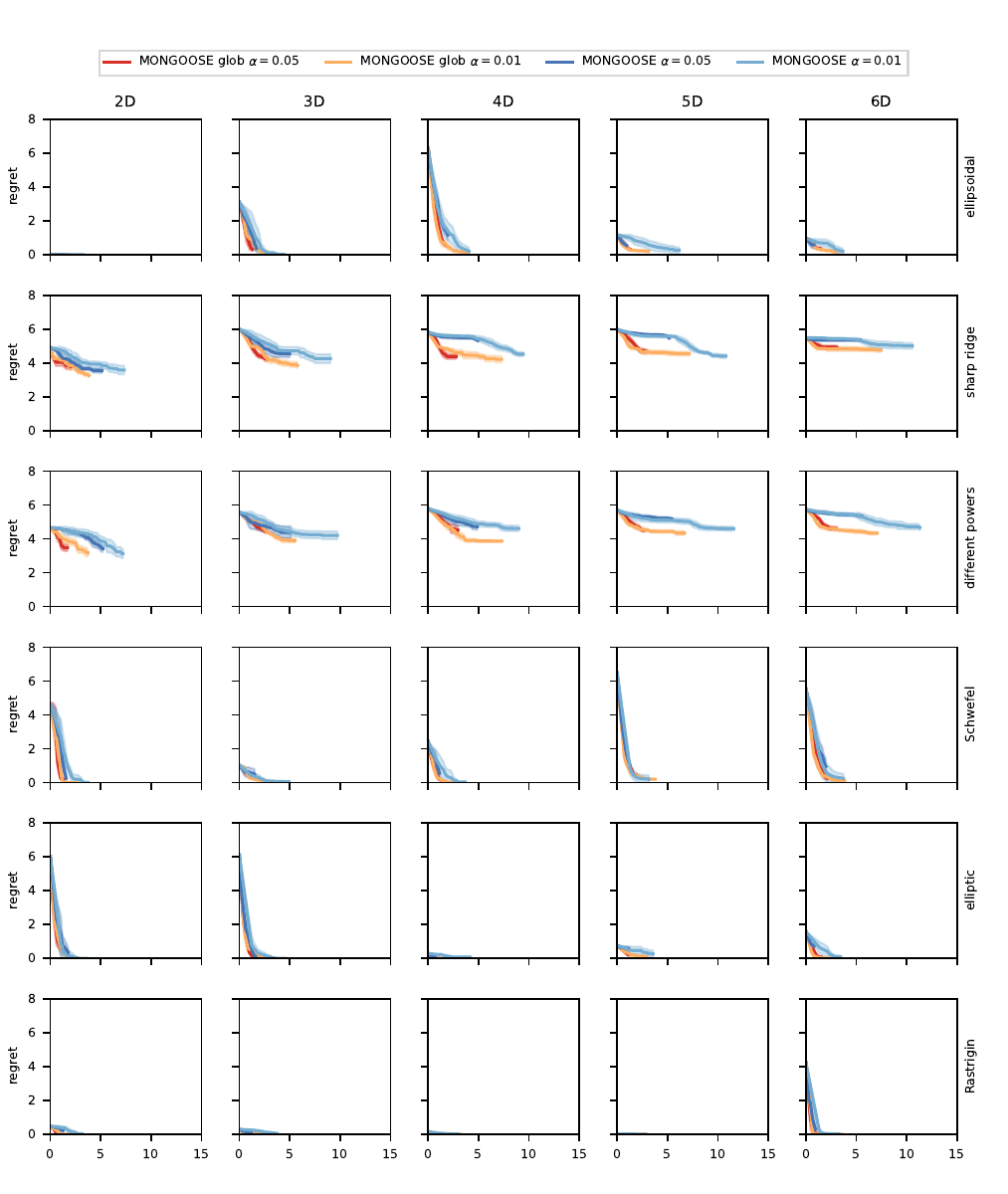}
    \caption{Individual COCO plots for Figure~\ref{fig:globstruct_coco_avg}. COCO functions 7-12: attractive sector function, Rosenbrock (original) function, Rosenbrock (rotated) function, ellipsoidal (non-separable) function, discus function, bent cigar function.}
    \label{fig:coco_50_2_glob}
\end{figure}

\begin{figure}
    \centering
    \includegraphics{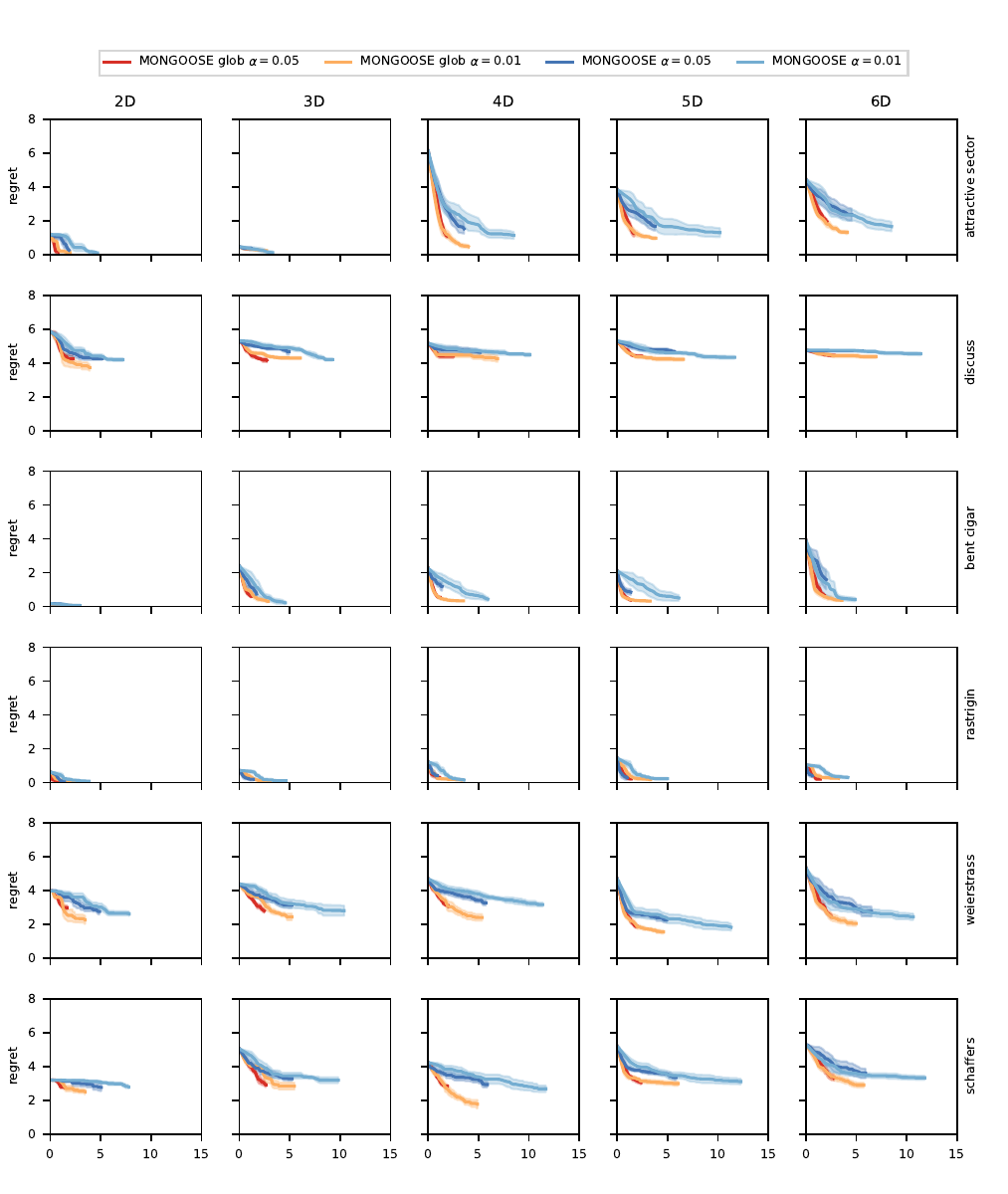}
    \caption{Individual COCO plots for Figure~\ref{fig:globstruct_coco_avg}. COCO functions 13-18: sharp ridge function, different powers function, Rastrigin (non-separable) function, Weierstrrass function, Schaffers F7 function, Schaffers F7 (moderately ill-conditioned) function.}
    \label{fig:coco_50_3_glob}
\end{figure}

\begin{figure}
    \centering
    \includegraphics{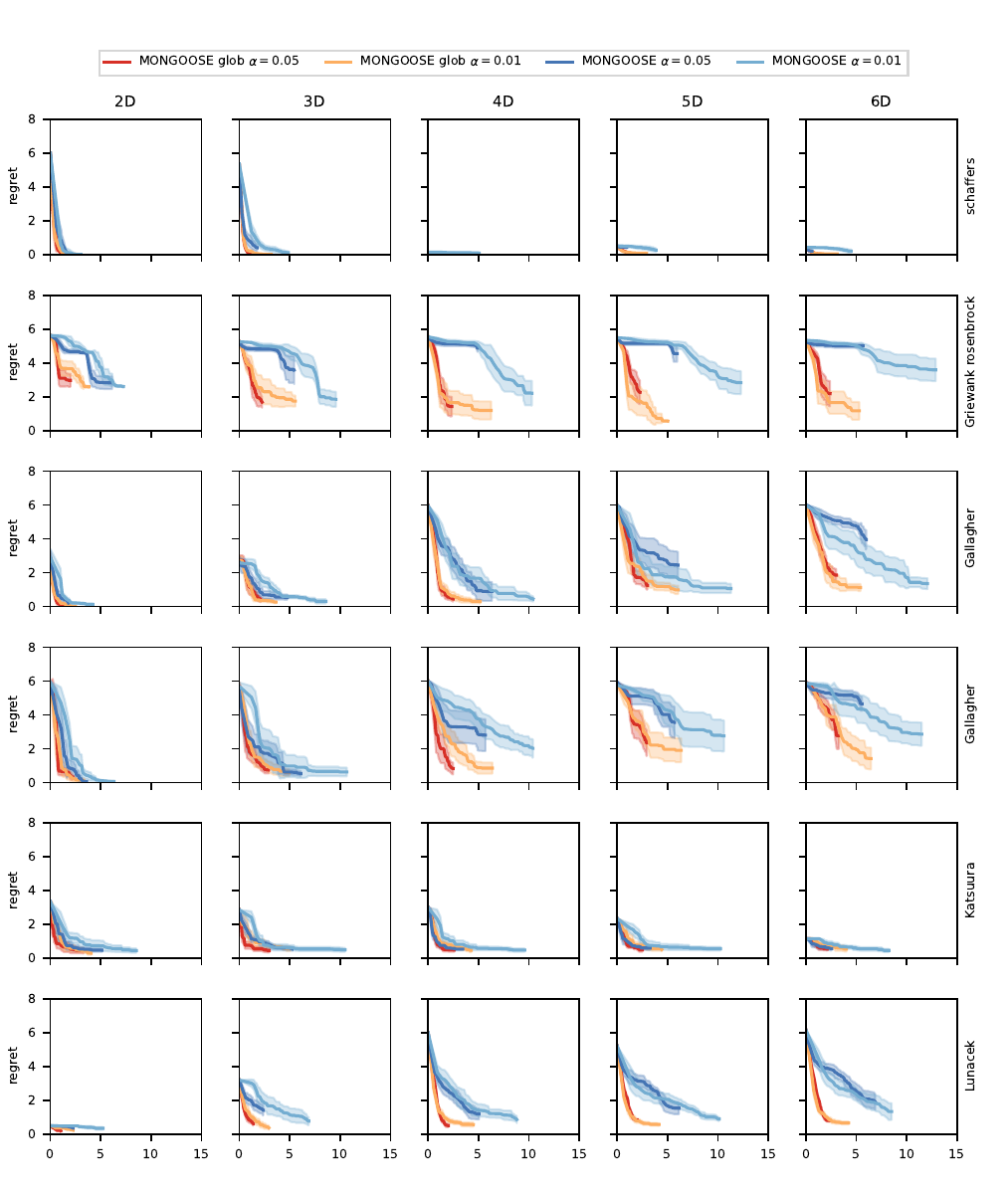}
    \caption{Individual COCO plots for Figure~\ref{fig:globstruct_coco_avg}. COCO functions 19-24: composite Griewank-Rosenbrock function, Schwefel function, Gallagher's Gaussian 101-me peaks function, Gallagher's Gaussian 21-hi peaks function, Kastsuura function, Lunacek bi-Rastrigin function.}
    \label{fig:coco_50_4_glob}
\end{figure}

\begin{figure}
    \centering
    \includegraphics{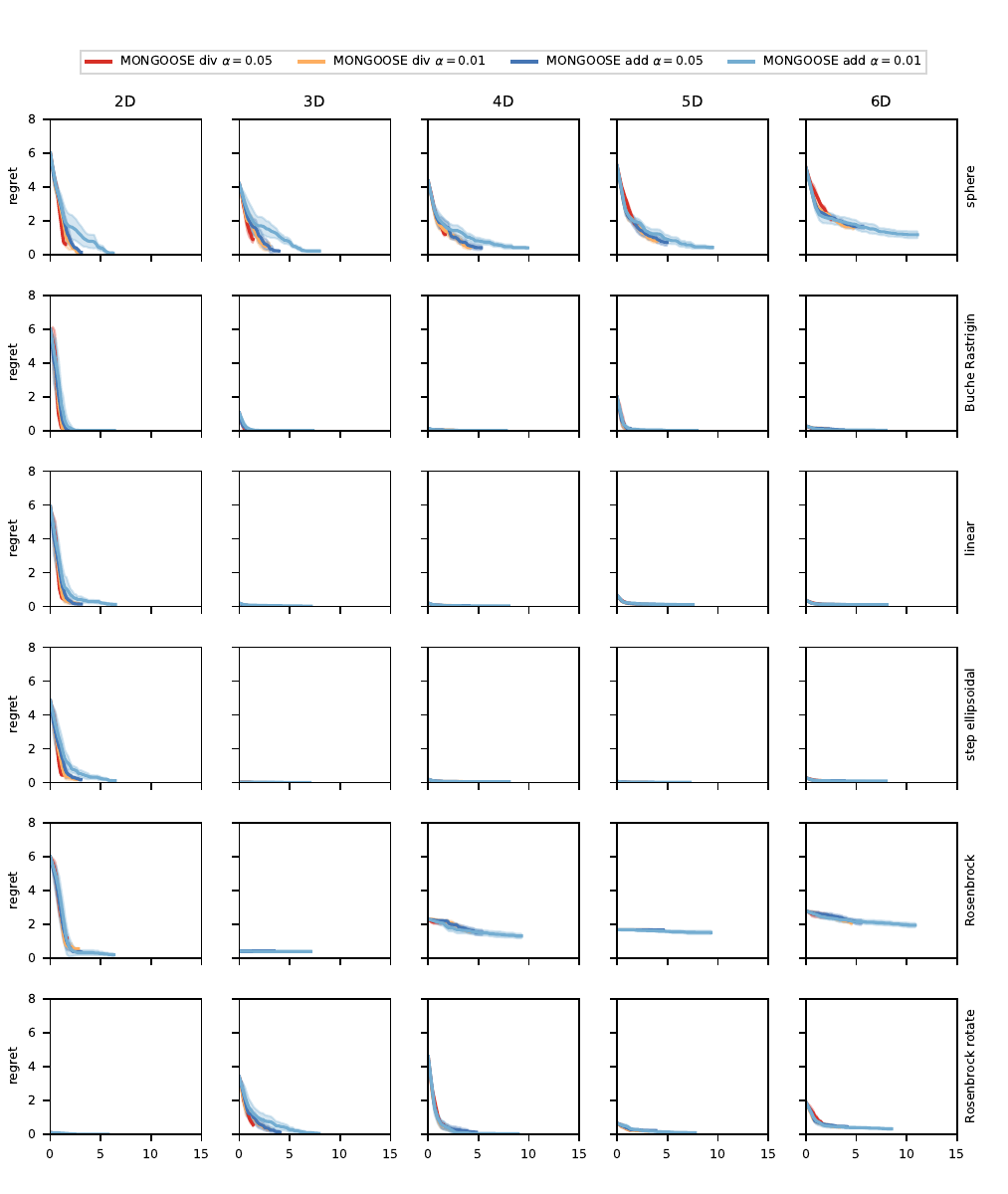}
    \caption{Individual COCO plots for Figure~\ref{fig:divadd_coco}. COCO functions 1-6: sphere function, ellipsoidal function, Rastrigin function, B\"{u}che-Rastrigin function, linear slope, step ellipsoidal function.}
    \label{fig:coco_50_1_divadd}
\end{figure}

\begin{figure}
    \centering
    \includegraphics{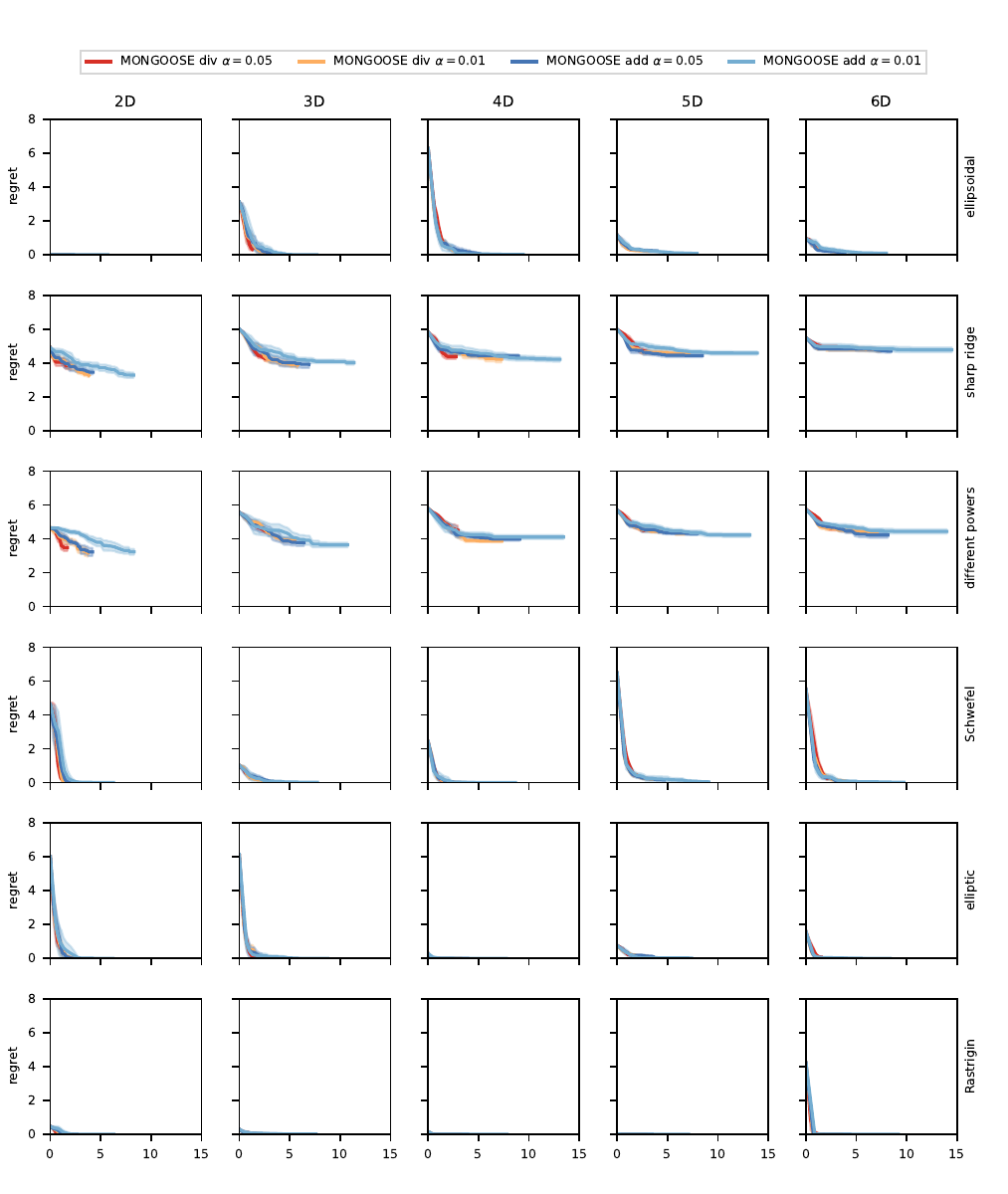}
    \caption{Individual COCO plots for Figure~\ref{fig:divadd_coco}. COCO functions 7-12: attractive sector function, Rosenbrock (original) function, Rosenbrock (rotated) function, ellipsoidal (non-separable) function, discus function, bent cigar function.}
    \label{fig:coco_50_2_divadd}
\end{figure}

\begin{figure}
    \centering
    \includegraphics{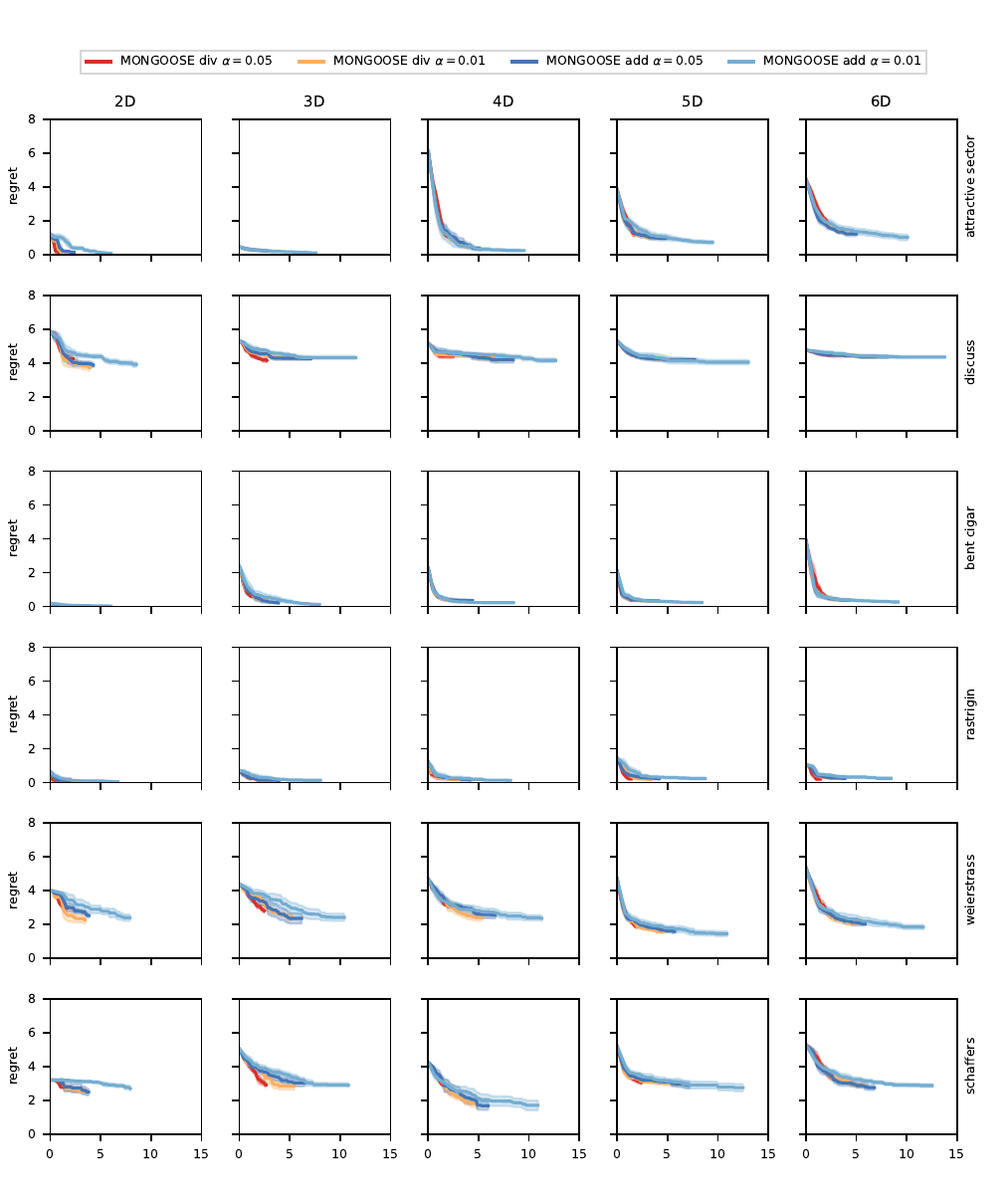}
    \caption{Individual COCO plots for Figure~\ref{fig:divadd_coco}. COCO functions 13-18: sharp ridge function, different powers function, Rastrigin (non-separable) function, Weierstrrass function, Schaffers F7 function, Schaffers F7 (moderately ill-conditioned) function.}
    \label{fig:coco_50_3_divadd}
\end{figure}

\begin{figure}
    \centering
    \includegraphics{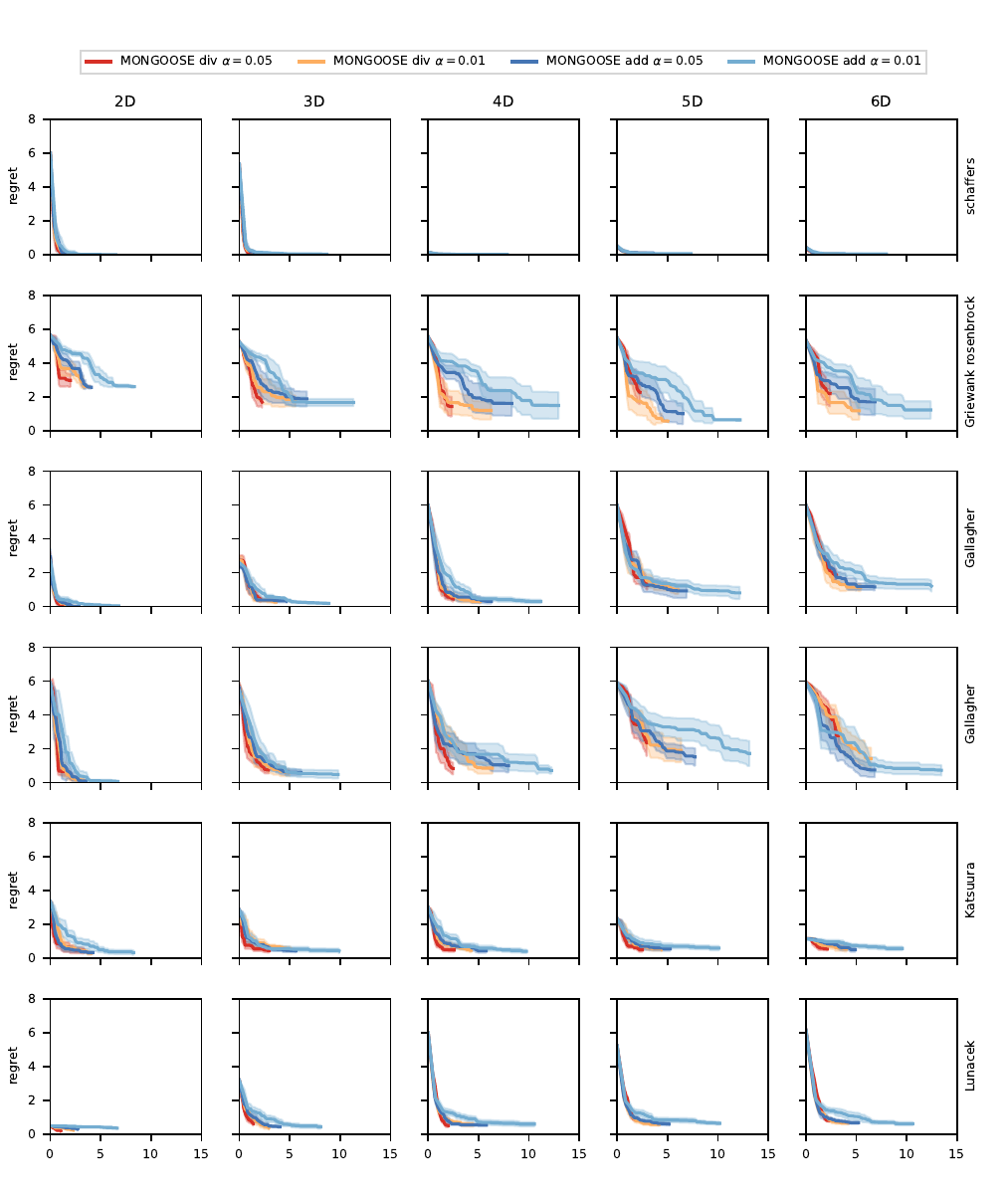}
    \caption{Individual COCO plots for Figure~\ref{fig:divadd_coco}. COCO functions 19-24: composite Griewank-Rosenbrock function, Schwefel function, Gallagher's Gaussian 101-me peaks function, Gallagher's Gaussian 21-hi peaks function, Kastsuura function, Lunacek bi-Rastrigin function.}
    \label{fig:coco_50_4_divadd}
\end{figure}

\begin{figure}
    \centering
    \includegraphics{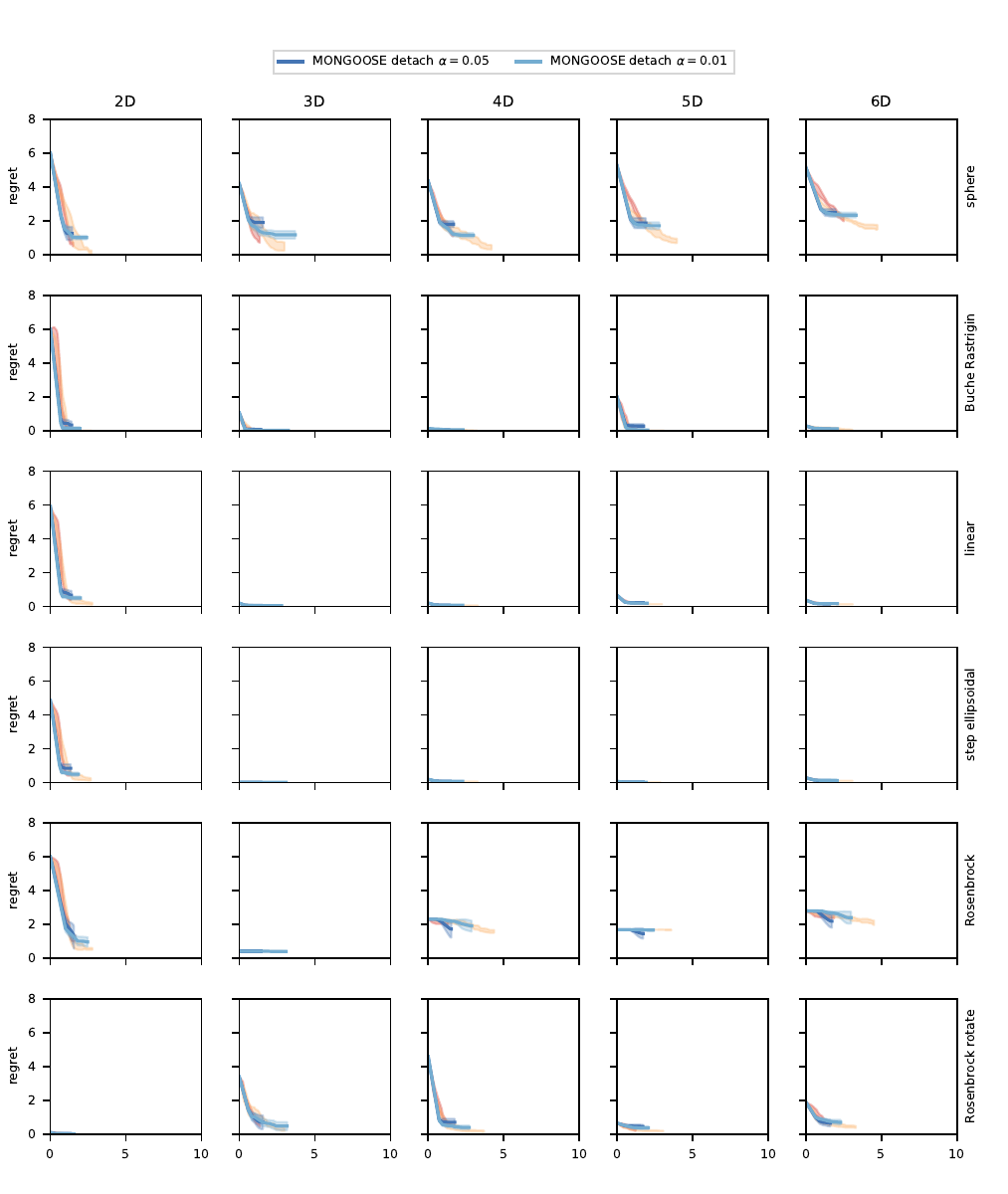}
    \caption{Individual COCO plots for Figure~\ref{fig:divadd_coco}. COCO functions 1-6: sphere function, ellipsoidal function, Rastrigin function, B\"{u}che-Rastrigin function, linear slope, step ellipsoidal function.}
    \label{fig:coco_detach_1}
\end{figure}

\begin{figure}
    \centering
    \includegraphics{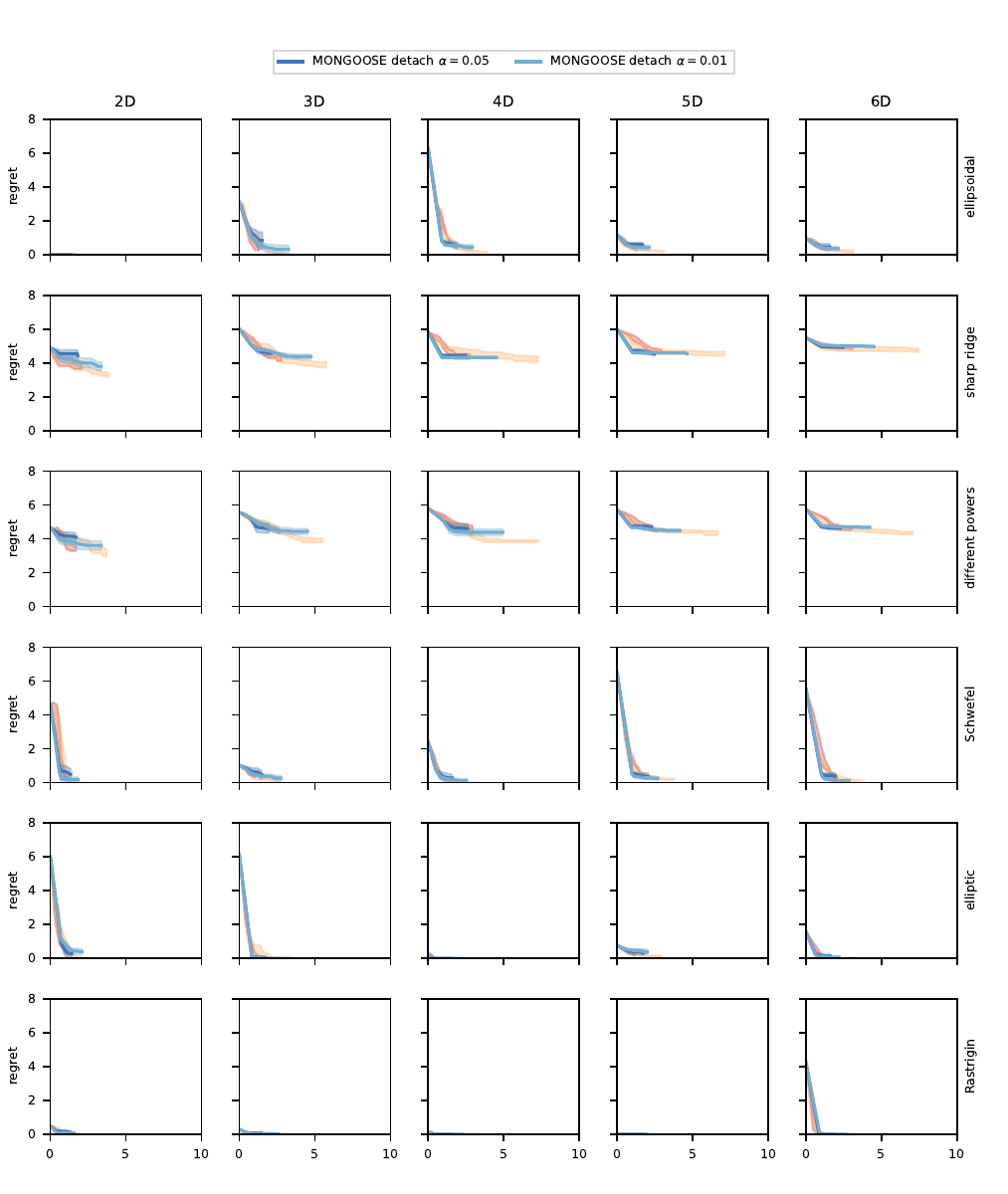}
    \caption{Individual COCO plots for Figure~\ref{fig:divadd_coco}. COCO functions 7-12: attractive sector function, Rosenbrock (original) function, Rosenbrock (rotated) function, ellipsoidal (non-separable) function, discus function, bent cigar function.}
    \label{fig:coco_detach_2}
\end{figure}

\begin{figure}
    \centering
    \includegraphics{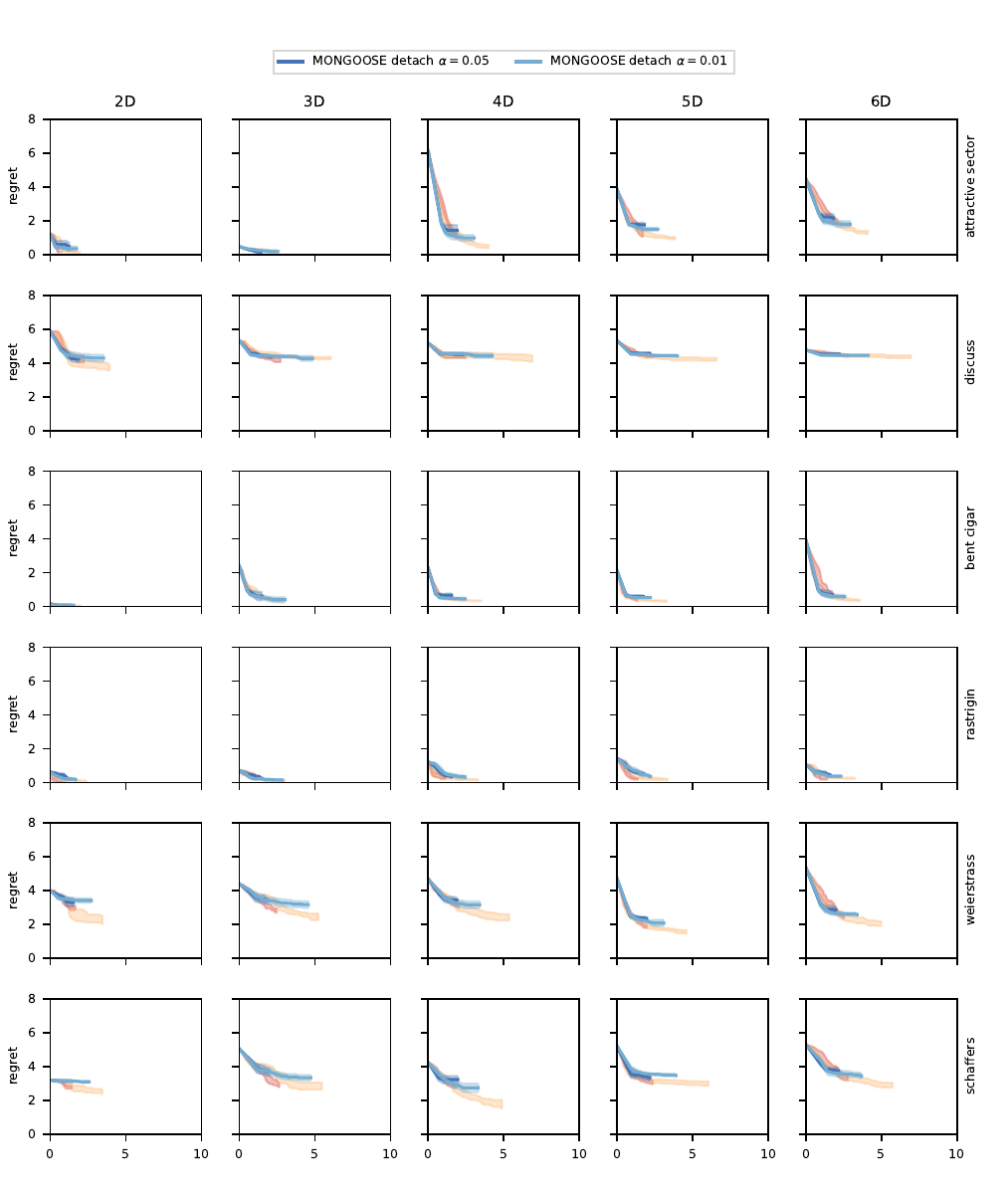}
    \caption{Individual COCO plots for Figure~\ref{fig:divadd_coco}. COCO functions 13-18: sharp ridge function, different powers function, Rastrigin (non-separable) function, Weierstrrass function, Schaffers F7 function, Schaffers F7 (moderately ill-conditioned) function.}
    \label{fig:coco_detach_3}
\end{figure}

\begin{figure}
    \centering
    \includegraphics{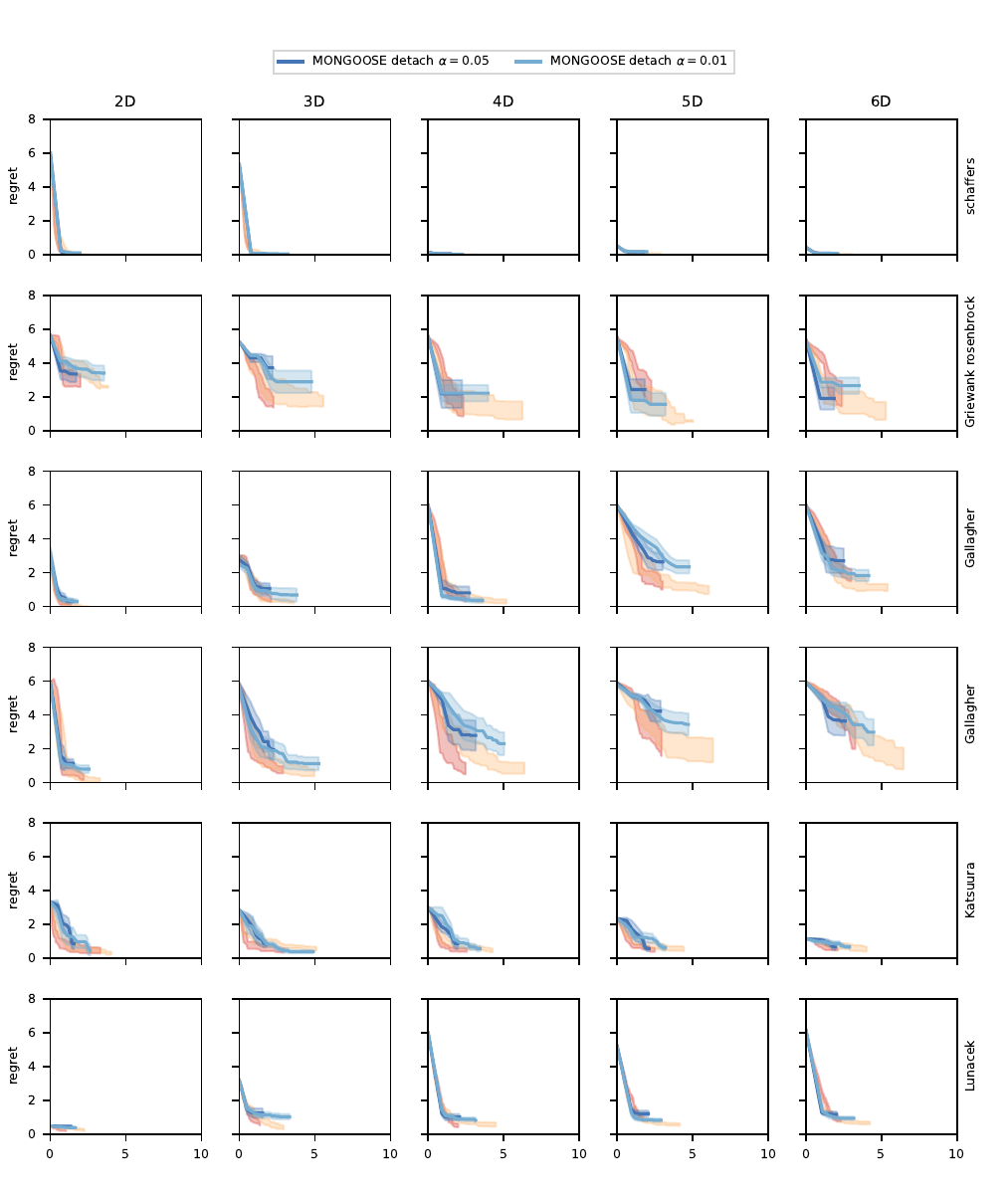}
    \caption{Individual COCO plots for Figure~\ref{fig:coco_detach}. COCO functions 19-24: composite Griewank-Rosenbrock function, Schwefel function, Gallagher's Gaussian 101-me peaks function, Gallagher's Gaussian 21-hi peaks function, Kastsuura function, Lunacek bi-Rastrigin function.}
    \label{fig:coco_detach_4}
\end{figure}

\begin{figure}
    \centering
    \includegraphics{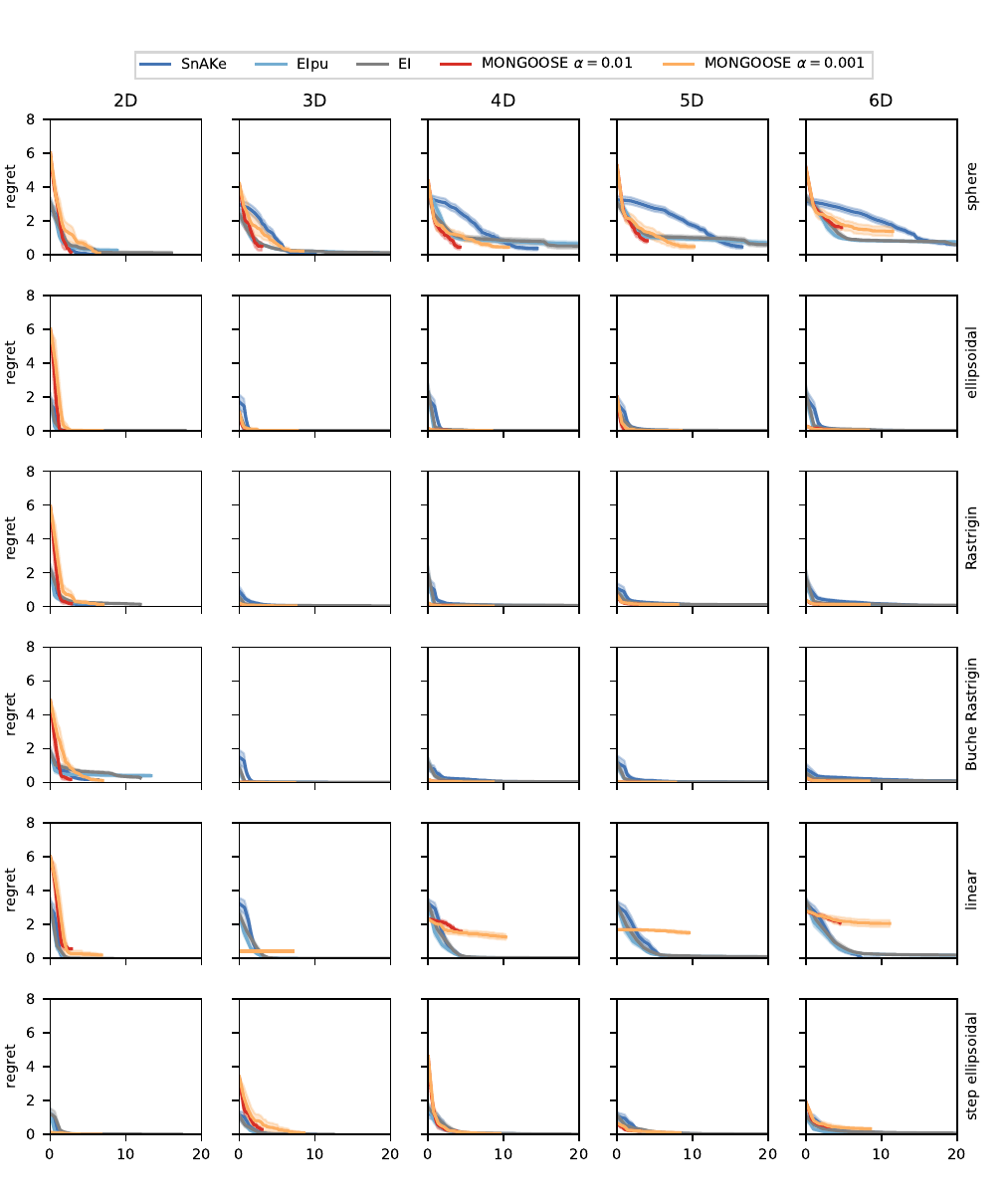}
    \caption{Individual COCO plots for Figure~\ref{fig:zero_coco}. COCO functions 1-6: sphere function, ellipsoidal function, Rastrigin function, B\"{u}che-Rastrigin function, linear slope, step ellipsoidal function.}
    \label{fig:coco_50_1_zero}
\end{figure}

\begin{figure}
    \centering
    \includegraphics{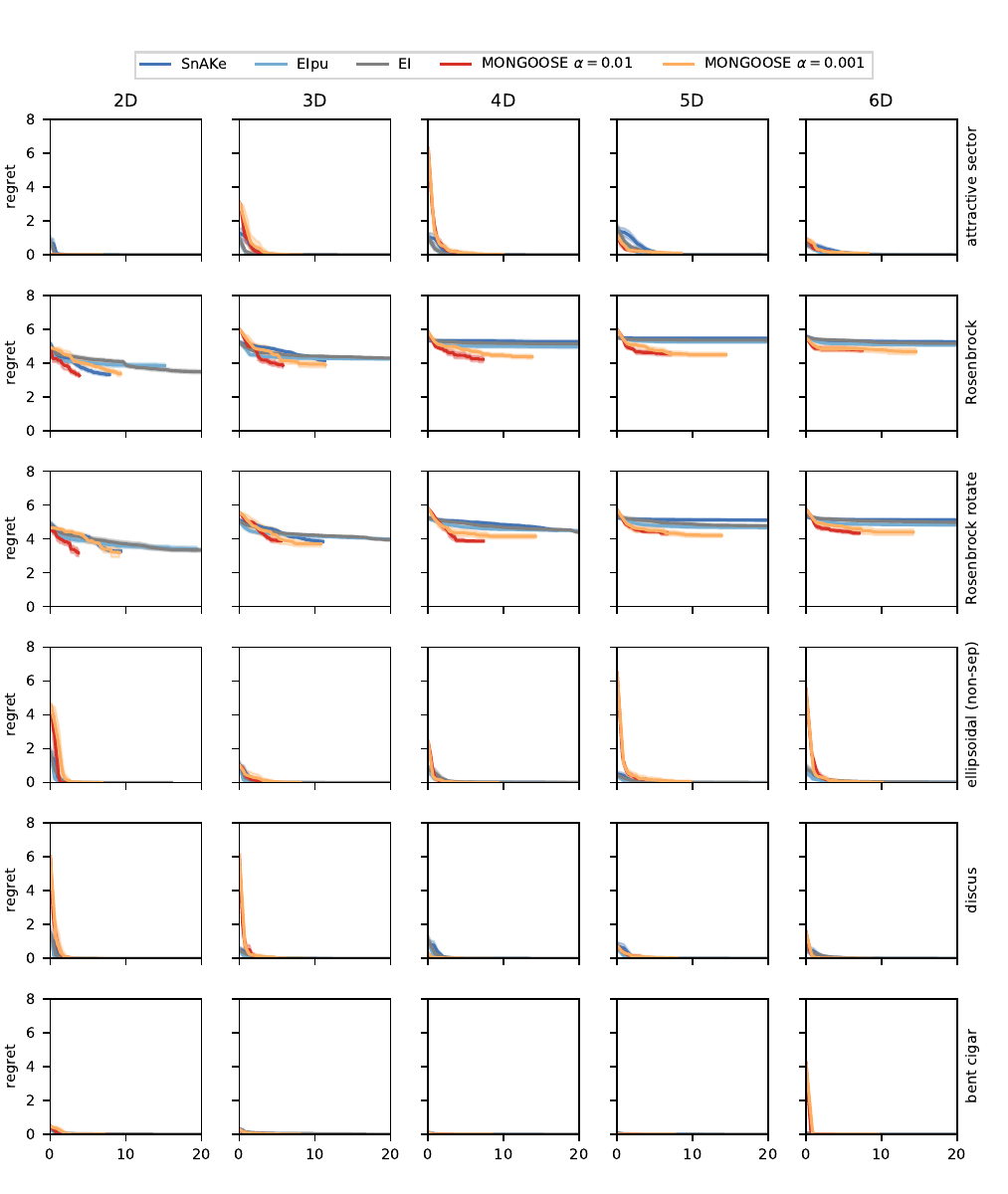}
    \caption{Individual COCO plots for Figure~\ref{fig:zero_coco}. COCO functions 7-12: attractive sector function, Rosenbrock (original) function, Rosenbrock (rotated) function, ellipsoidal (non-separable) function, discus function, bent cigar function.}
    \label{fig:coco_50_2_zero}
\end{figure}

\begin{figure}
    \centering
    \includegraphics{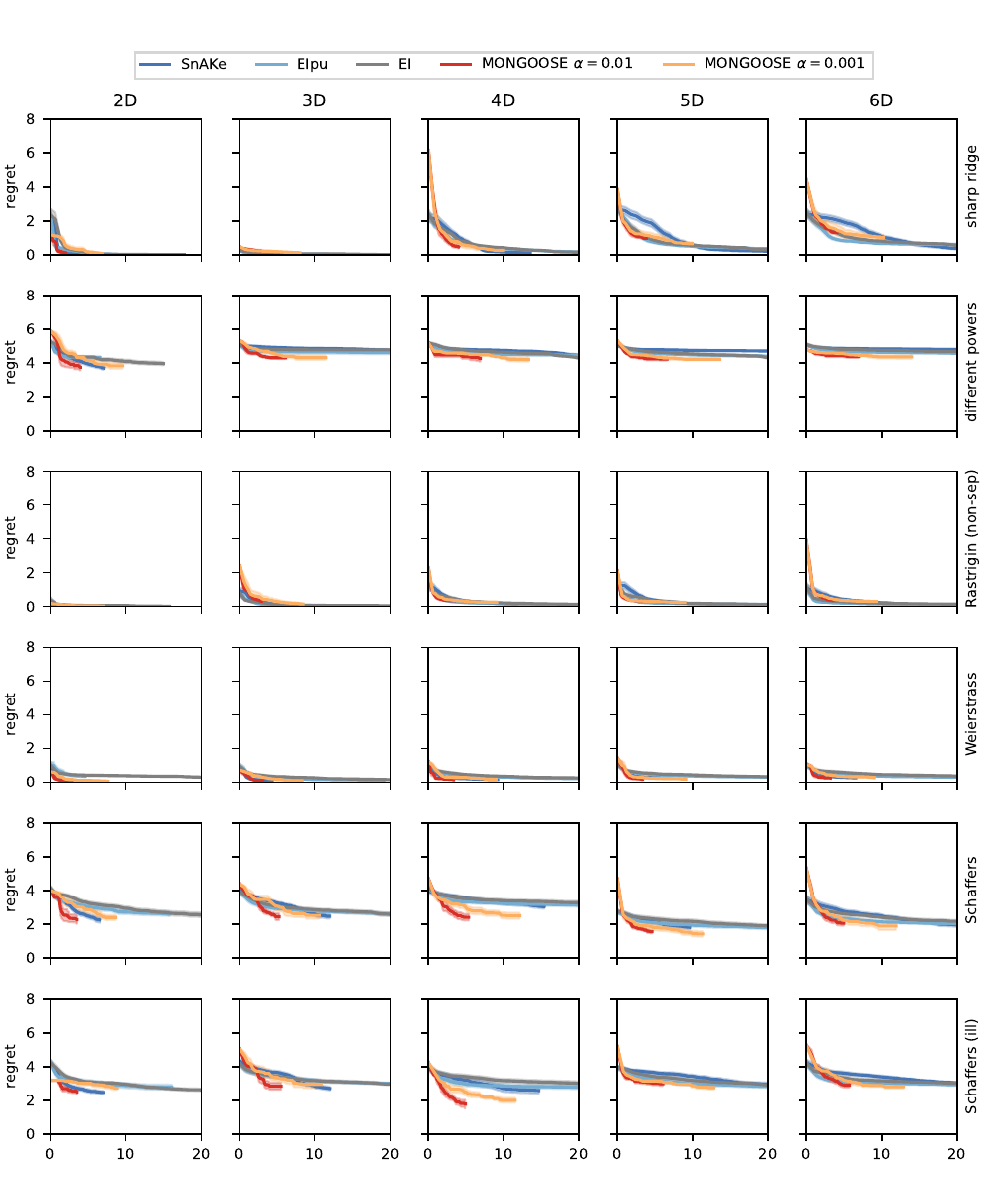}
    \caption{Individual COCO plots for Figure~\ref{fig:zero_coco}. COCO functions 13-18: sharp ridge function, different powers function, Rastrigin (non-separable) function, Weierstrrass function, Schaffers F7 function, Schaffers F7 (moderately ill-conditioned) function.}
    \label{fig:coco_50_3_zero}
\end{figure}

\begin{figure}
    \centering
    \includegraphics{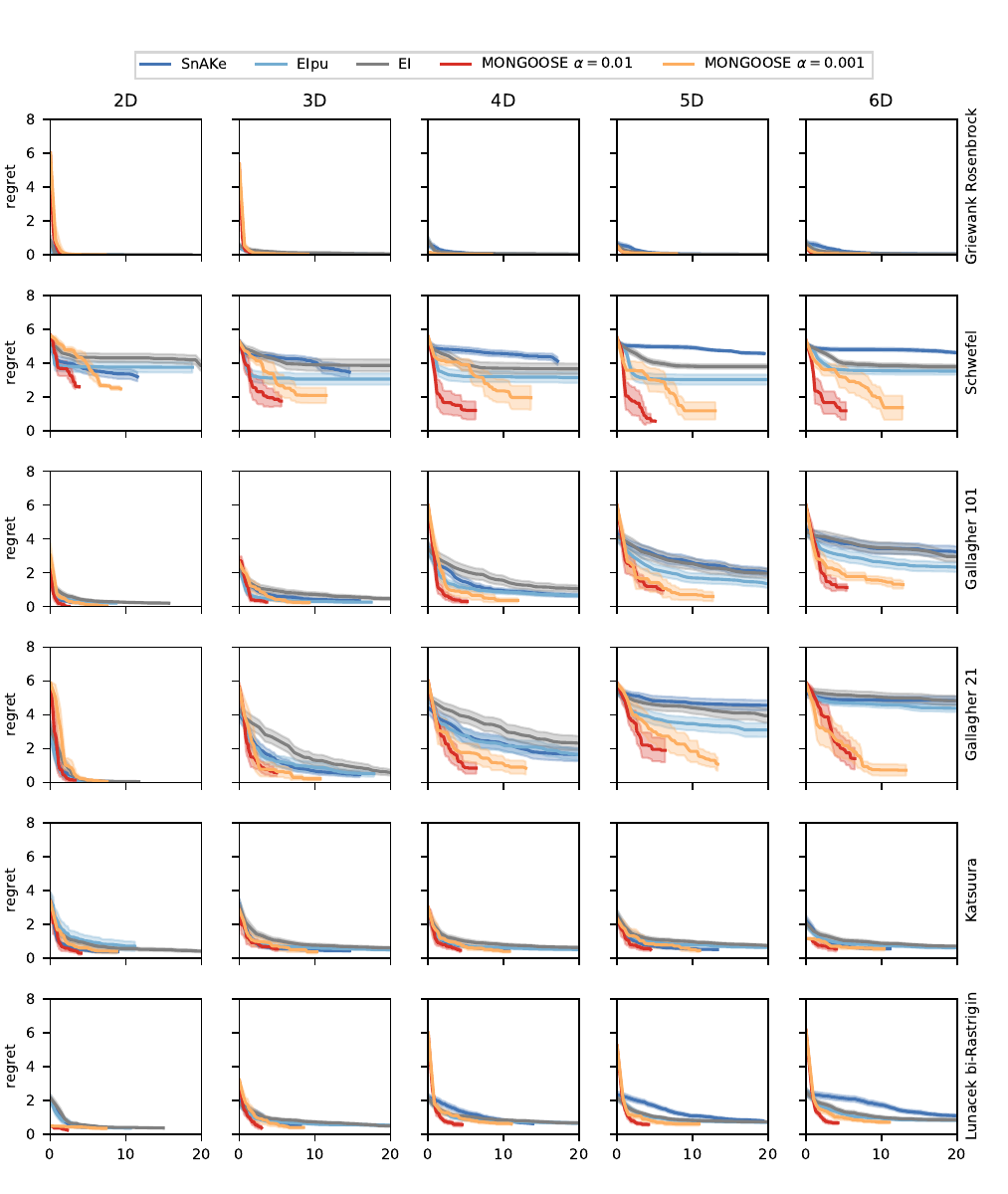}
    \caption{Individual COCO plots for Figure~\ref{fig:zero_coco}. COCO functions 19-24: composite Griewank-Rosenbrock function, Schwefel function, Gallagher's Gaussian 101-me peaks function, Gallagher's Gaussian 21-hi peaks function, Kastsuura function, Lunacek bi-Rastrigin function.}
    \label{fig:coco_50_4_zero}
\end{figure}

\begin{figure}
    \centering
    \includegraphics{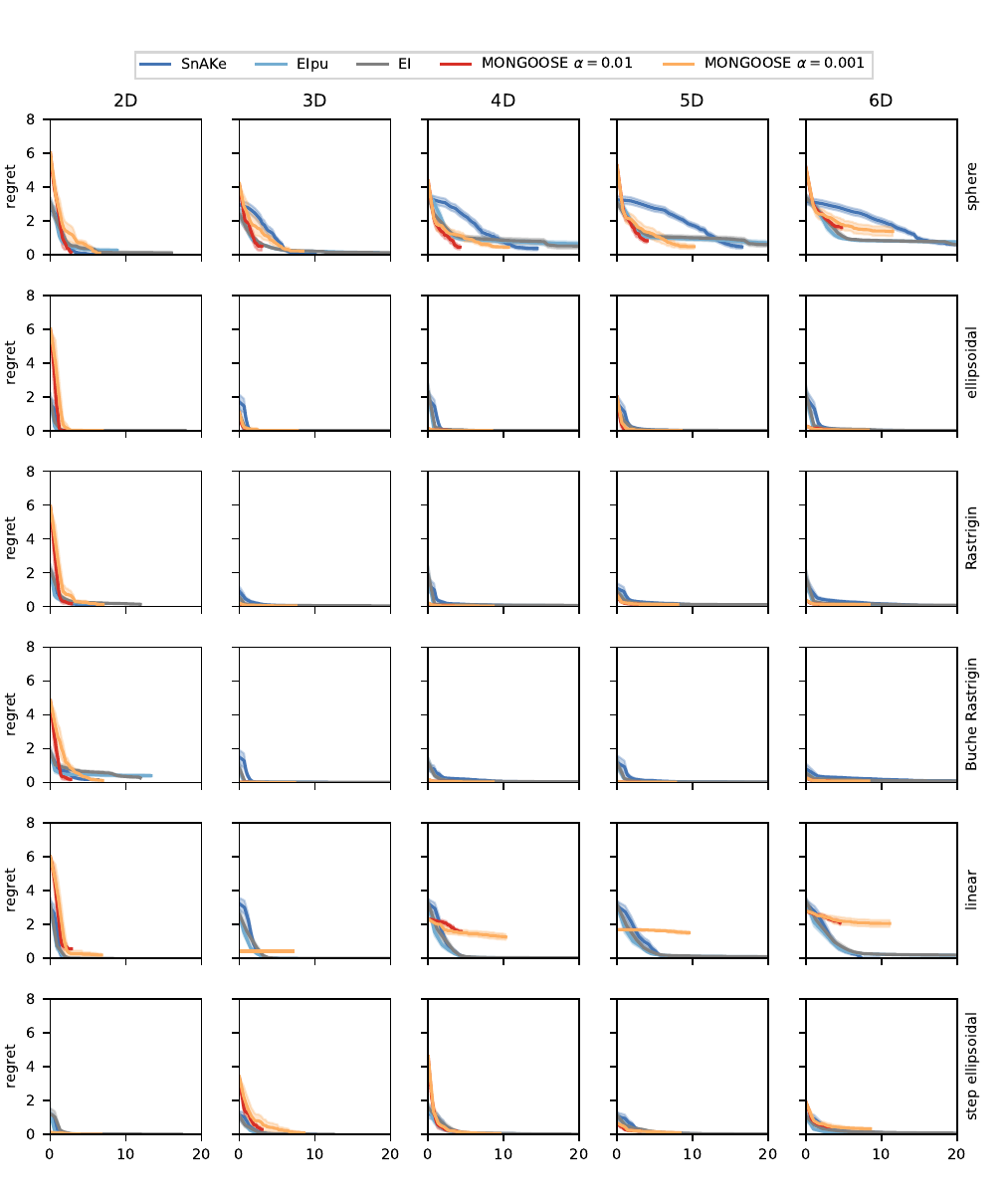}
    \caption{Individual COCO plots for Figure~\ref{fig:coco_50_1_noisy_0.1}. COCO functions 1-6: sphere function, ellipsoidal function, Rastrigin function, B\"{u}che-Rastrigin function, linear slope, step ellipsoidal function.}
    \label{fig:coco_50_1_noisy_0.1}
\end{figure}

\begin{figure}
    \centering
    \includegraphics{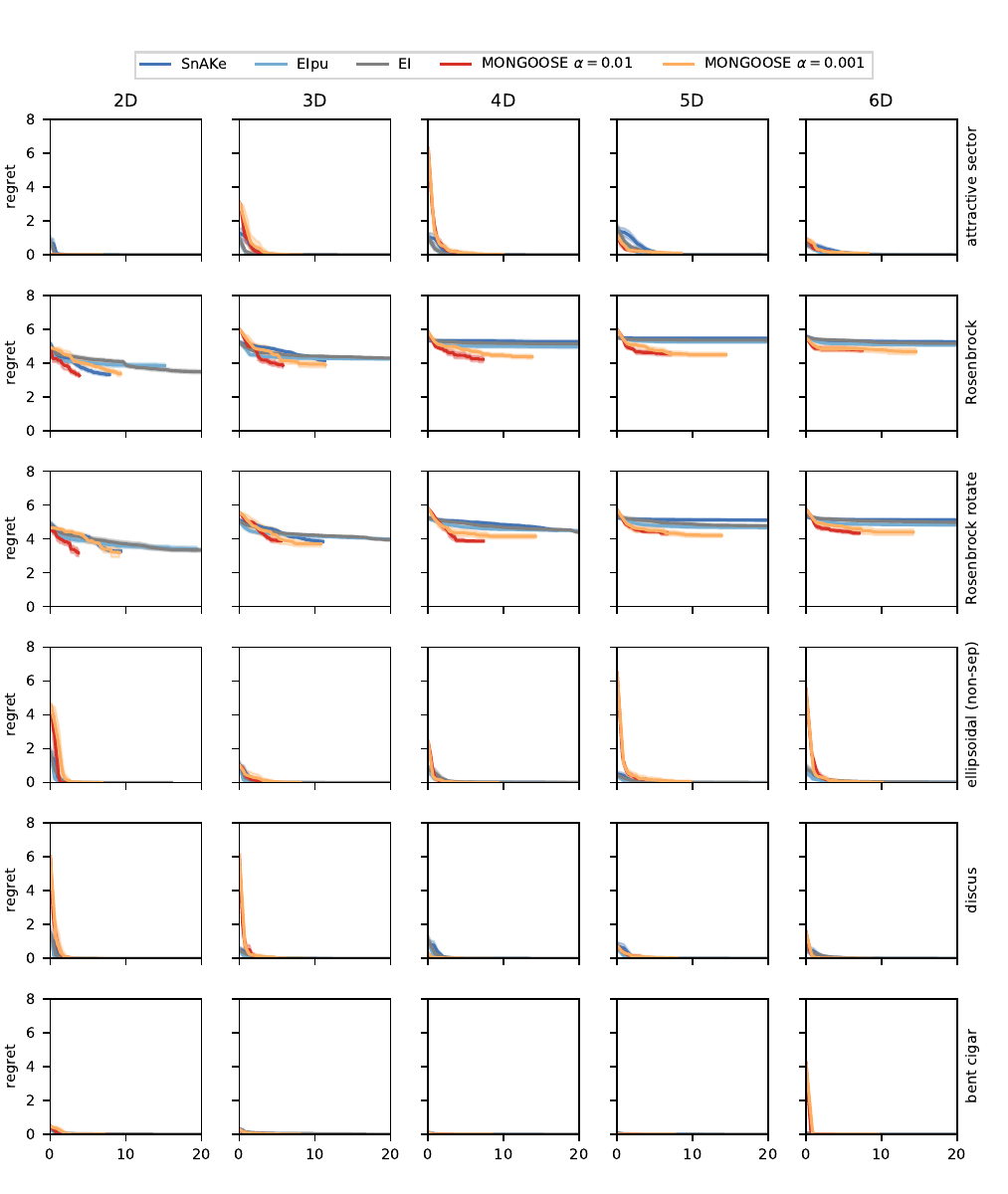}
    \caption{Individual COCO plots for Figure~\ref{fig:coco_50_1_noisy_0.1}. COCO functions 7-12: attractive sector function, Rosenbrock (original) function, Rosenbrock (rotated) function, ellipsoidal (non-separable) function, discus function, bent cigar function.}
    \label{fig:coco_50_2_noisy_0.1}
\end{figure}

\begin{figure}
    \centering
    \includegraphics{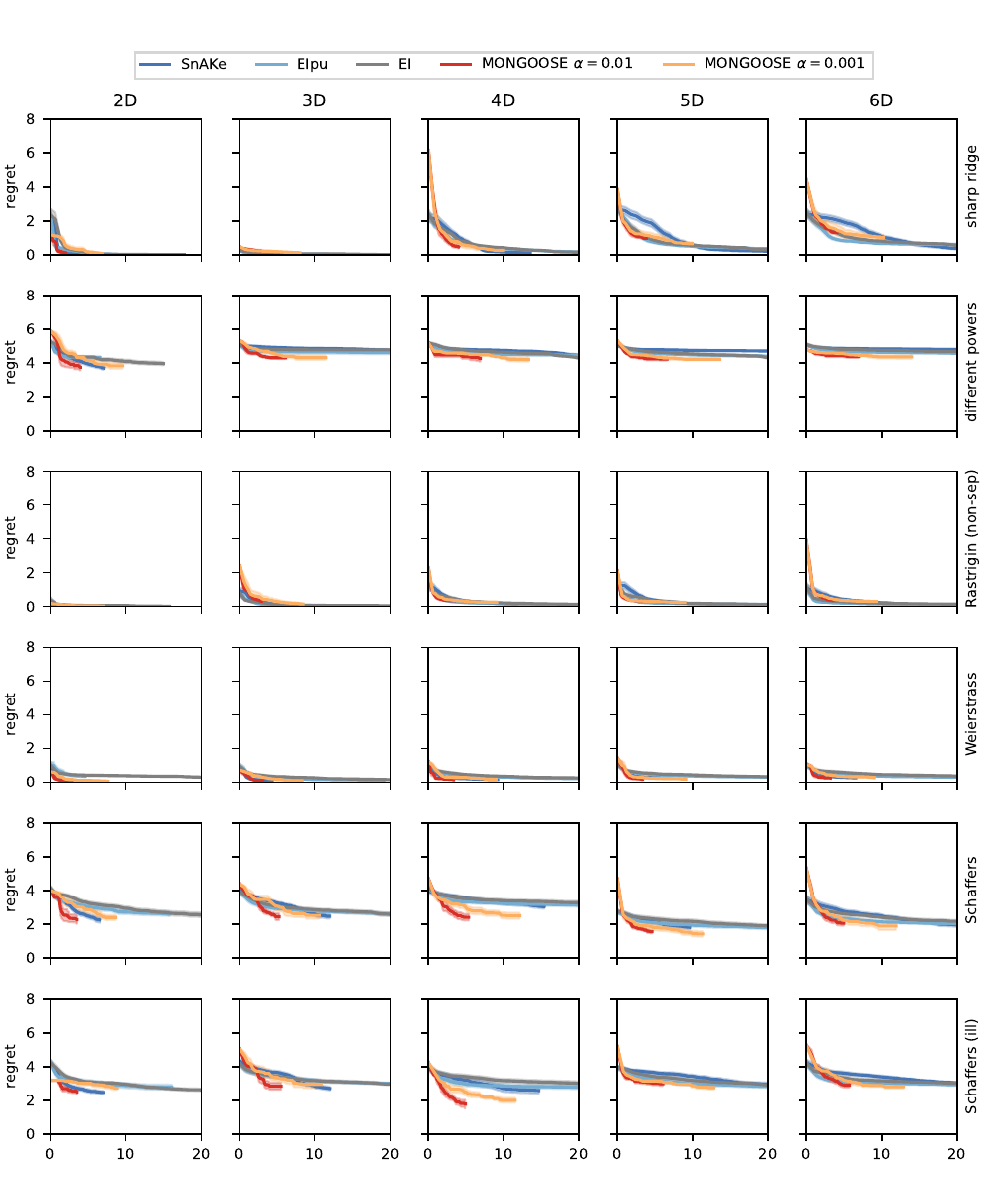}
    \caption{Individual COCO plots for Figure~\ref{fig:coco_50_1_noisy_0.1}. COCO functions 13-18: sharp ridge function, different powers function, Rastrigin (non-separable) function, Weierstrrass function, Schaffers F7 function, Schaffers F7 (moderately ill-conditioned) function.}
    \label{fig:coco_50_3_noisy_0.1}
\end{figure}

\begin{figure}
    \centering
    \includegraphics{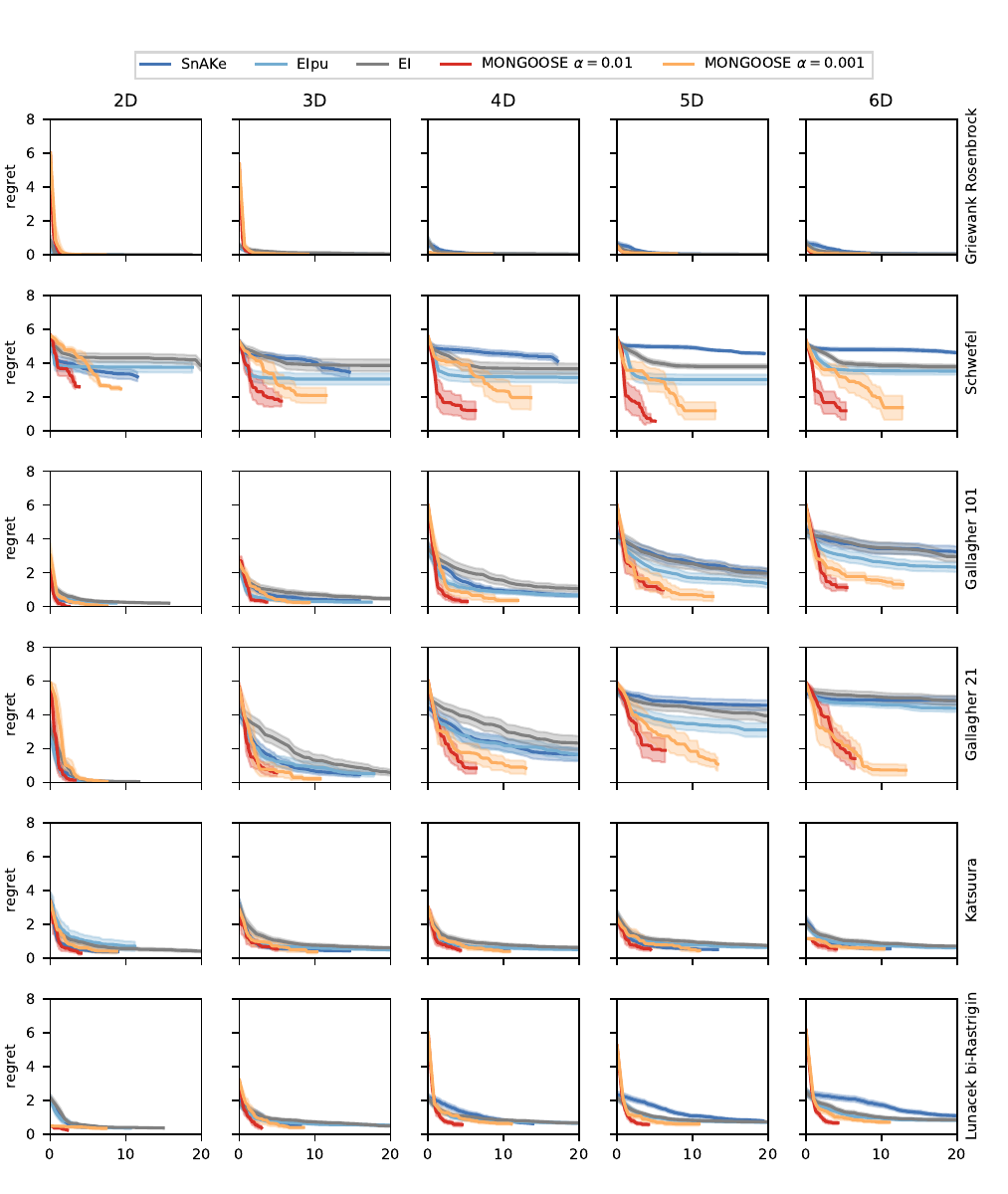}
    \caption{Individual COCO plots for Figure~\ref{fig:coco_50_1_noisy_0.1}. COCO functions 19-24: composite Griewank-Rosenbrock function, Schwefel function, Gallagher's Gaussian 101-me peaks function, Gallagher's Gaussian 21-hi peaks function, Kastsuura function, Lunacek bi-Rastrigin function.}
    \label{fig:coco_50_4_noisy_0.1}
\end{figure}

\begin{figure}
    \centering
    \includegraphics{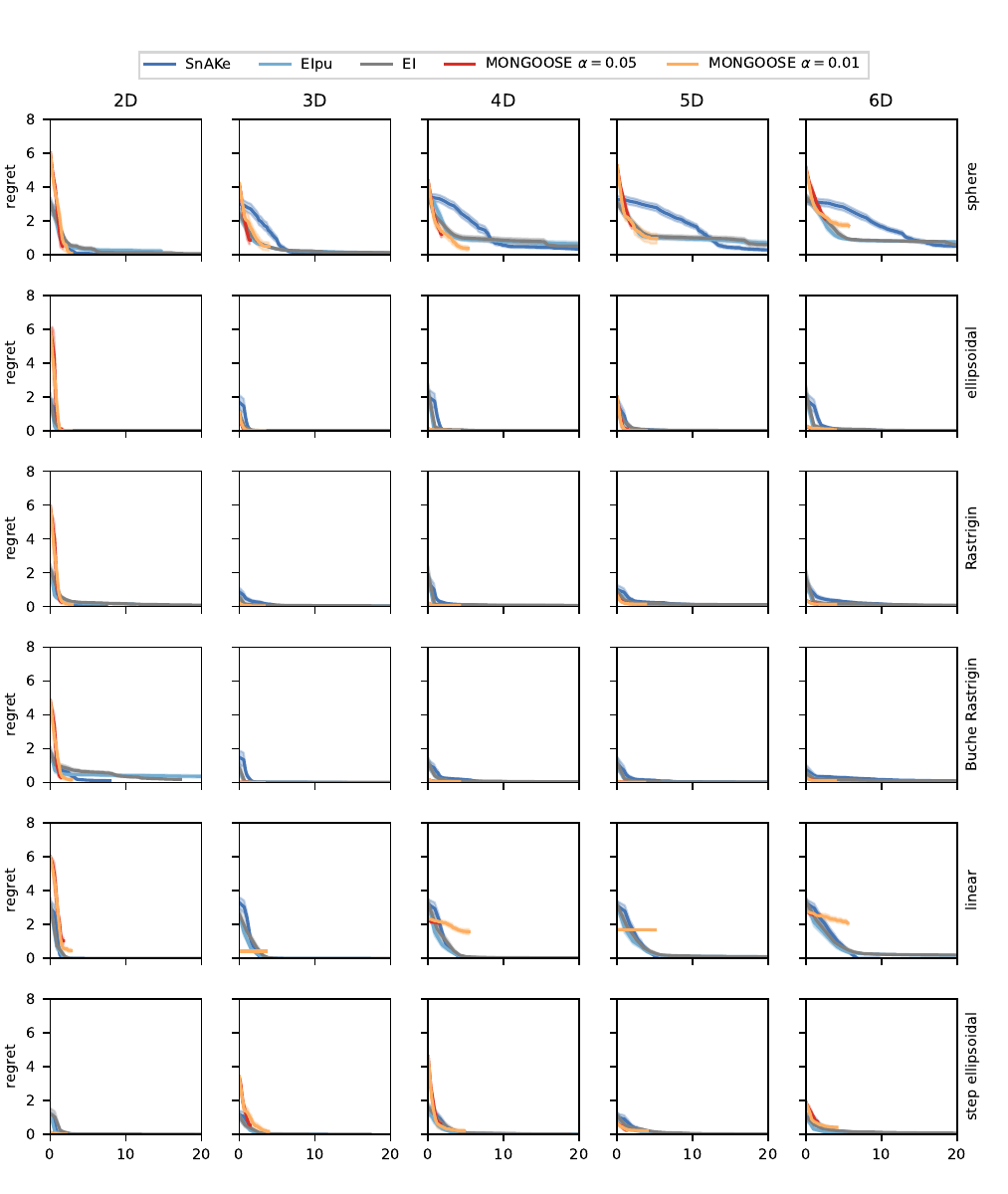}
    \caption{Individual COCO plots for Figure~\ref{fig:coco_100}. COCO functions 1-6: sphere function, ellipsoidal function, Rastrigin function, B\"{u}che-Rastrigin function, linear slope, step ellipsoidal function.}
    \label{fig:coco_100_1}
\end{figure}

\begin{figure}
    \centering
    \includegraphics{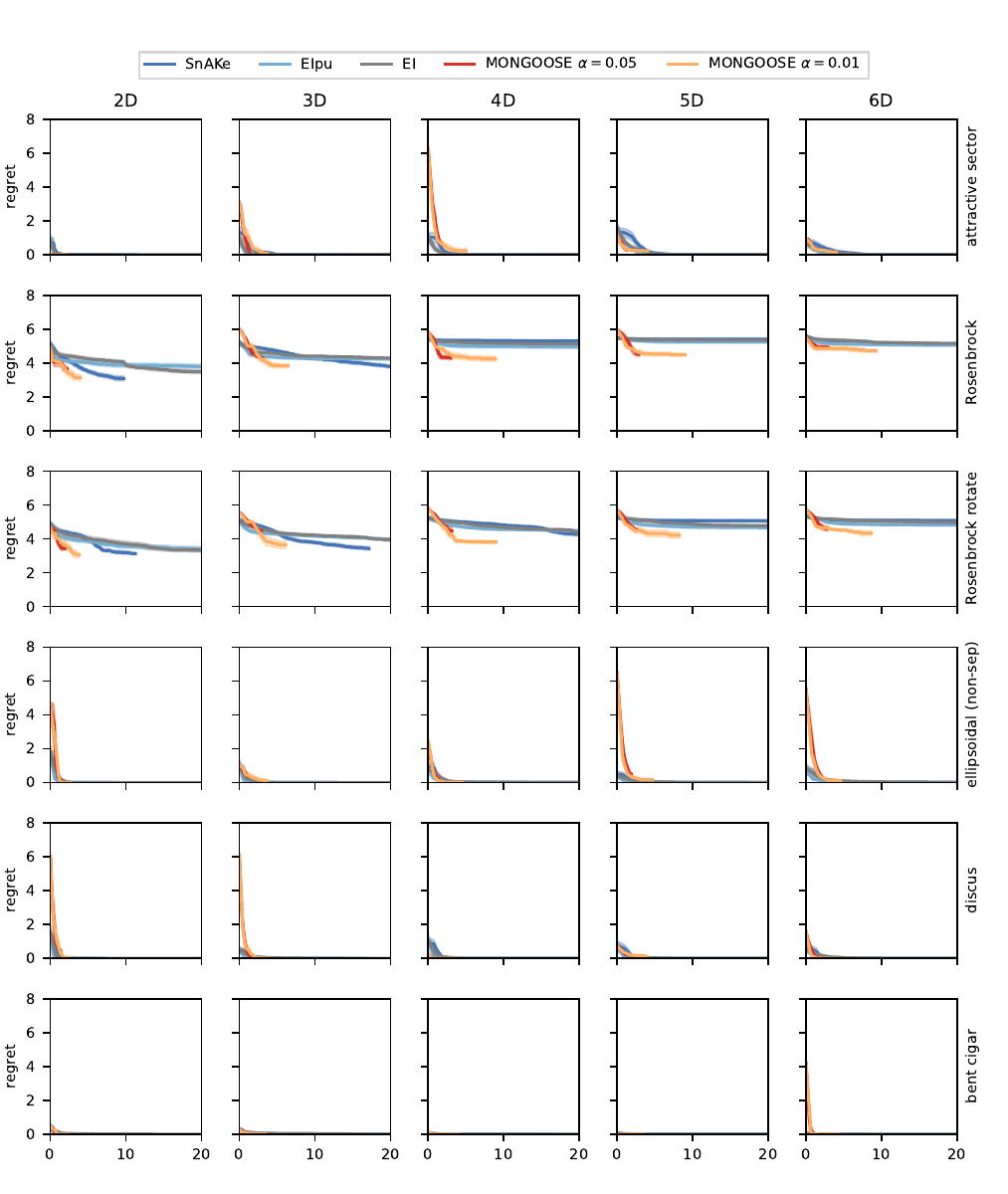}
    \caption{Individual COCO plots for Figure~\ref{fig:coco_100}. COCO functions 7-12: attractive sector function, Rosenbrock (original) function, Rosenbrock (rotated) function, ellipsoidal (non-separable) function, discus function, bent cigar function.}
    \label{fig:coco_100_2}
\end{figure}

\begin{figure}
    \centering
    \includegraphics{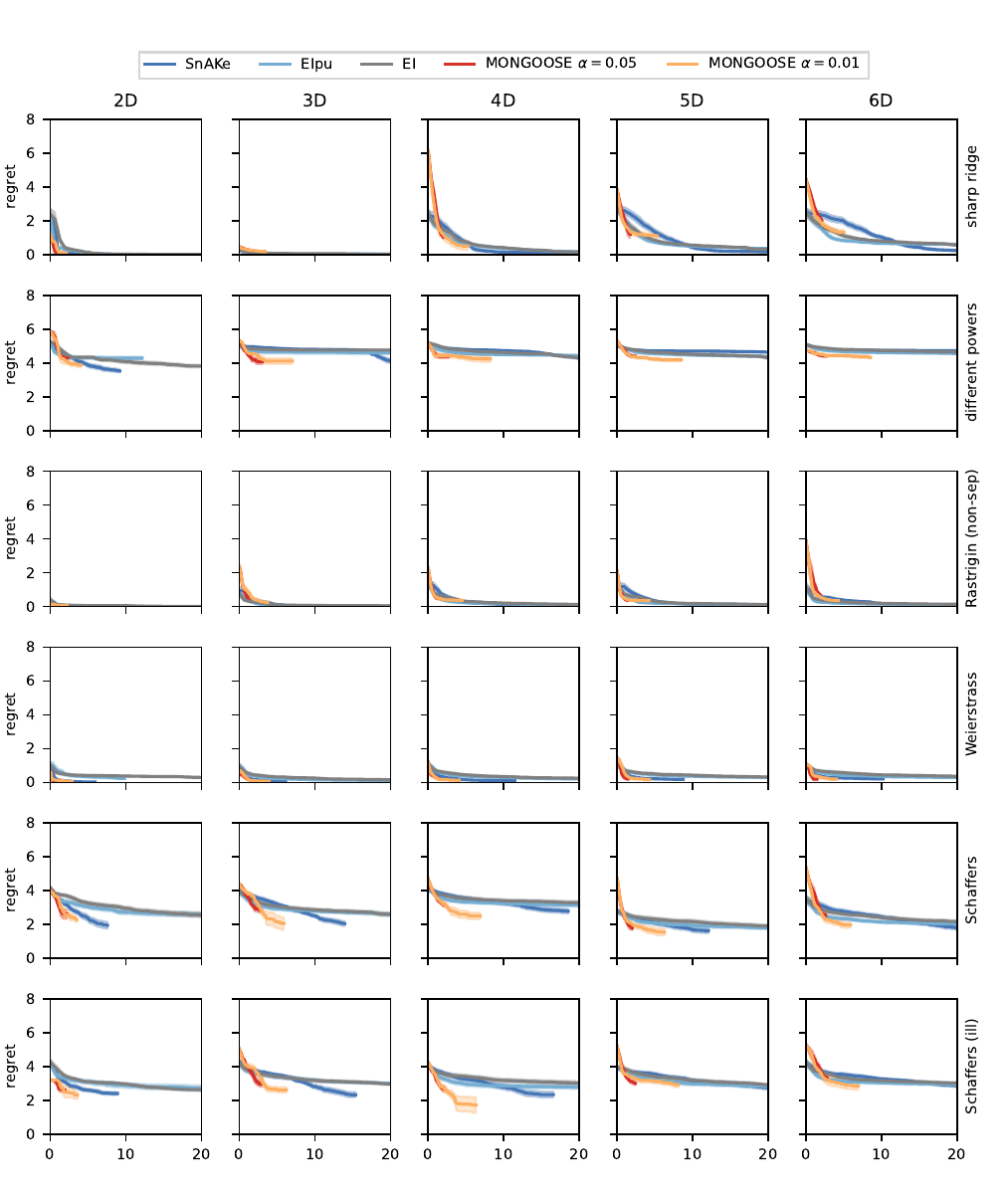}
    \caption{Individual COCO plots for Figure~\ref{fig:coco_100}. COCO functions 13-18: sharp ridge function, different powers function, Rastrigin (non-separable) function, Weierstrrass function, Schaffers F7 function, Schaffers F7 (moderately ill-conditioned) function.}
    \label{fig:coco_100_3}
\end{figure}

\begin{figure}
    \centering
    \includegraphics{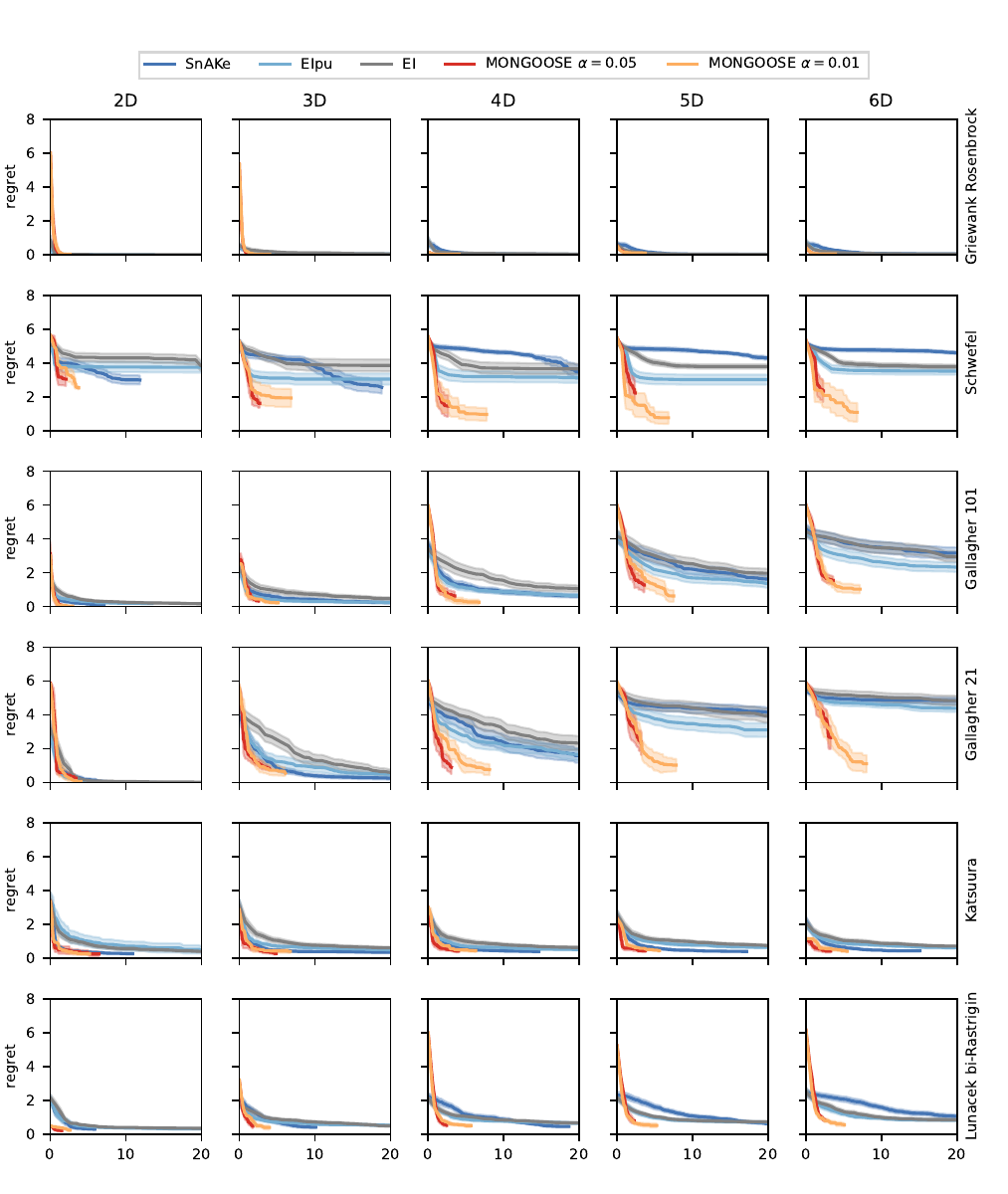}
    \caption{Individual COCO plots for Figure~\ref{fig:coco_100}. COCO functions 19-24: composite Griewank-Rosenbrock function, Schwefel function, Gallagher's Gaussian 101-me peaks function, Gallagher's Gaussian 21-hi peaks function, Kastsuura function, Lunacek bi-Rastrigin function.}
    \label{fig:coco_100_4}
\end{figure}

\begin{figure}
    \centering
    \includegraphics{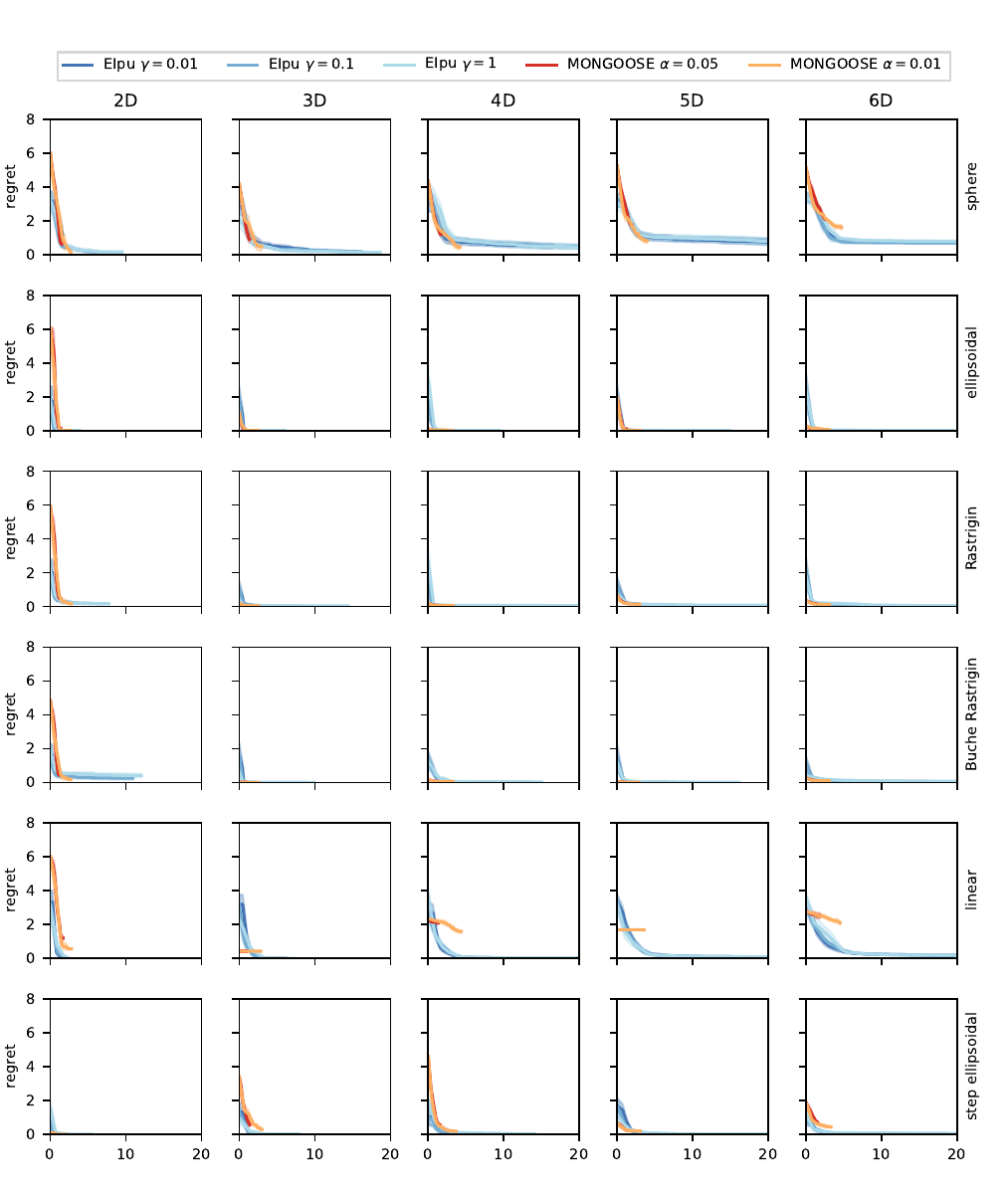}
    \caption{Individual COCO plots for Figure~\ref{fig:coco_eipu}. COCO functions 1-6: sphere function, ellipsoidal function, Rastrigin function, B\"{u}che-Rastrigin function, linear slope, step ellipsoidal function.}
    \label{fig:coco_eipu_1}
\end{figure}

\begin{figure}
    \centering
    \includegraphics{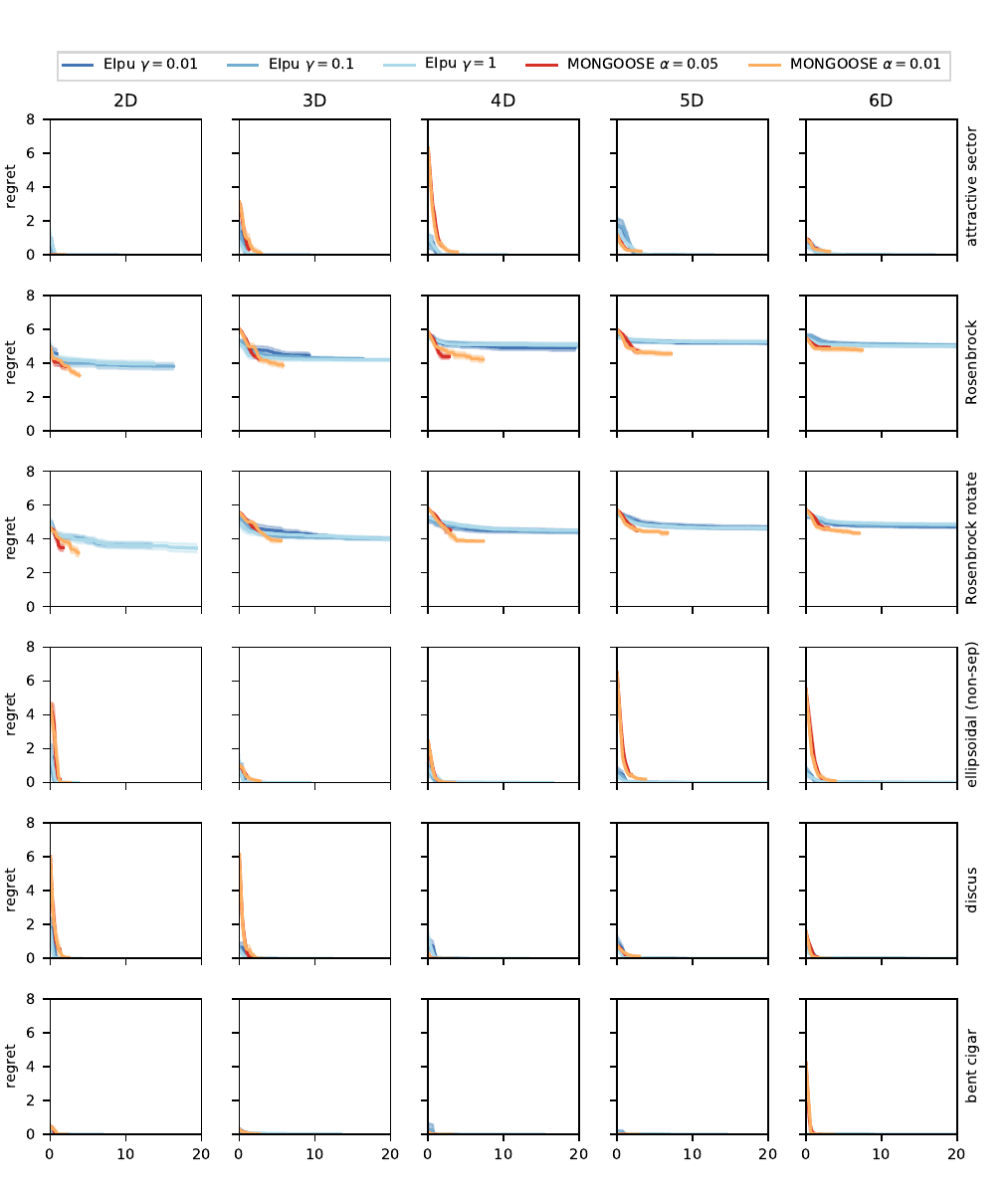}
    \caption{Individual COCO plots for Figure~\ref{fig:coco_eipu}. COCO functions 7-12: attractive sector function, Rosenbrock (original) function, Rosenbrock (rotated) function, ellipsoidal (non-separable) function, discus function, bent cigar function.}
    \label{fig:coco_eipu_2}
\end{figure}

\begin{figure}
    \centering
    \includegraphics{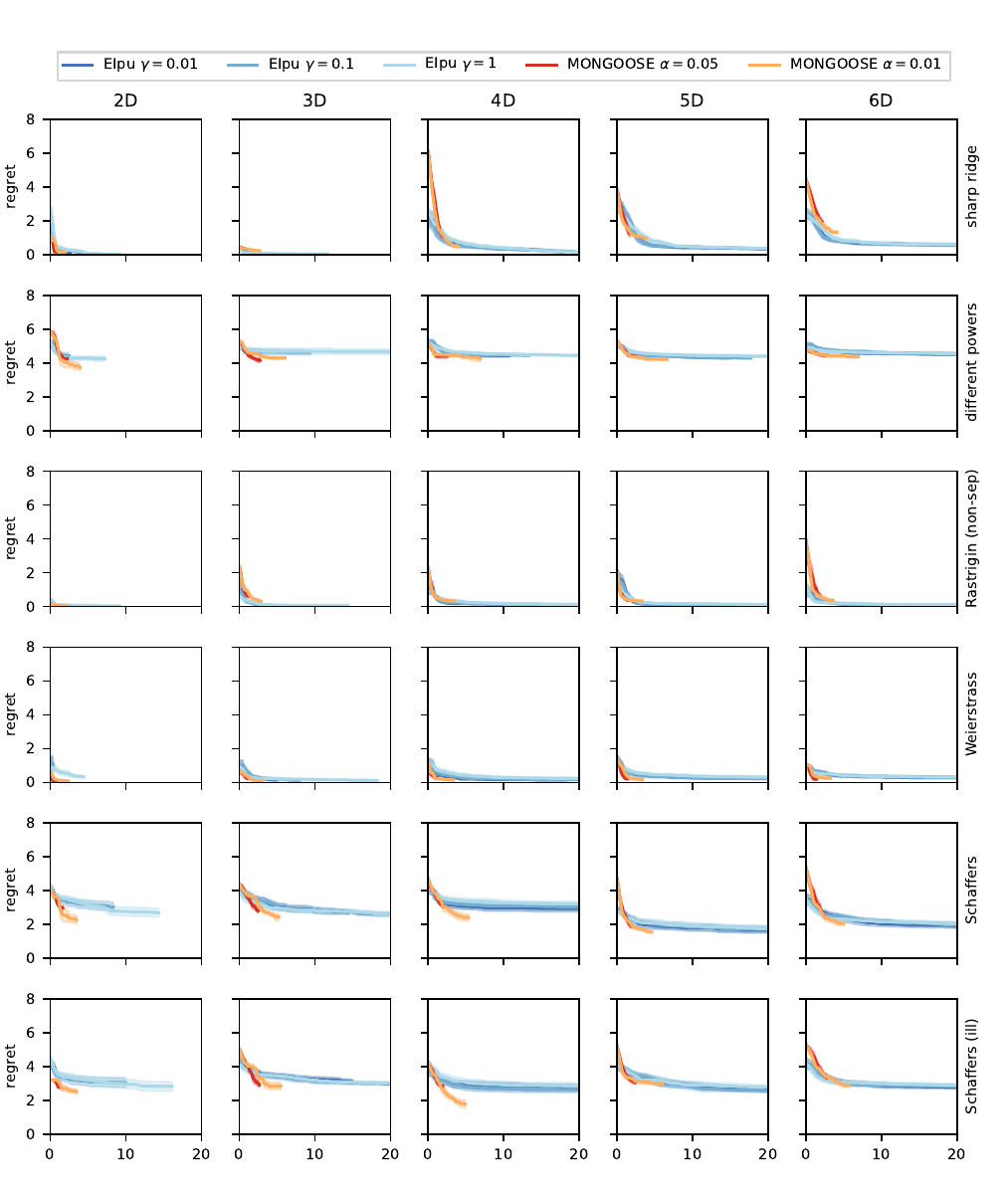}
    \caption{Individual COCO plots for Figure~\ref{fig:coco_eipu}. COCO functions 13-18: sharp ridge function, different powers function, Rastrigin (non-separable) function, Weierstrrass function, Schaffers F7 function, Schaffers F7 (moderately ill-conditioned) function.}
    \label{fig:coco_eipu_3}
\end{figure}

\begin{figure}
    \centering
    \includegraphics{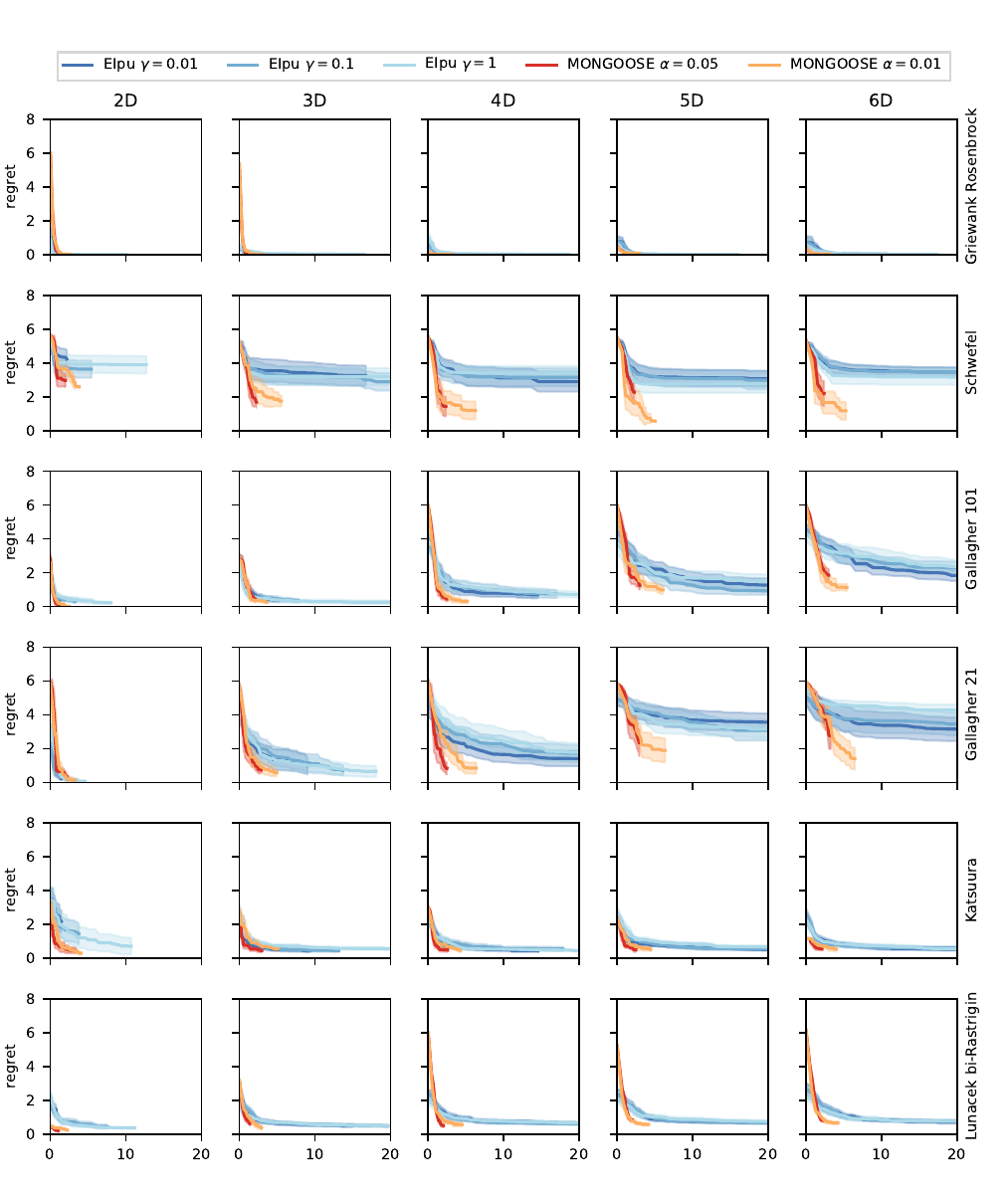}
    \caption{Individual COCO plots for Figure~\ref{fig:coco_eipu}. COCO functions 19-24: composite Griewank-Rosenbrock function, Schwefel function, Gallagher's Gaussian 101-me peaks function, Gallagher's Gaussian 21-hi peaks function, Kastsuura function, Lunacek bi-Rastrigin function.}
    \label{fig:coco_eipu_4}
\end{figure}

\begin{figure}
    \centering
    \includegraphics{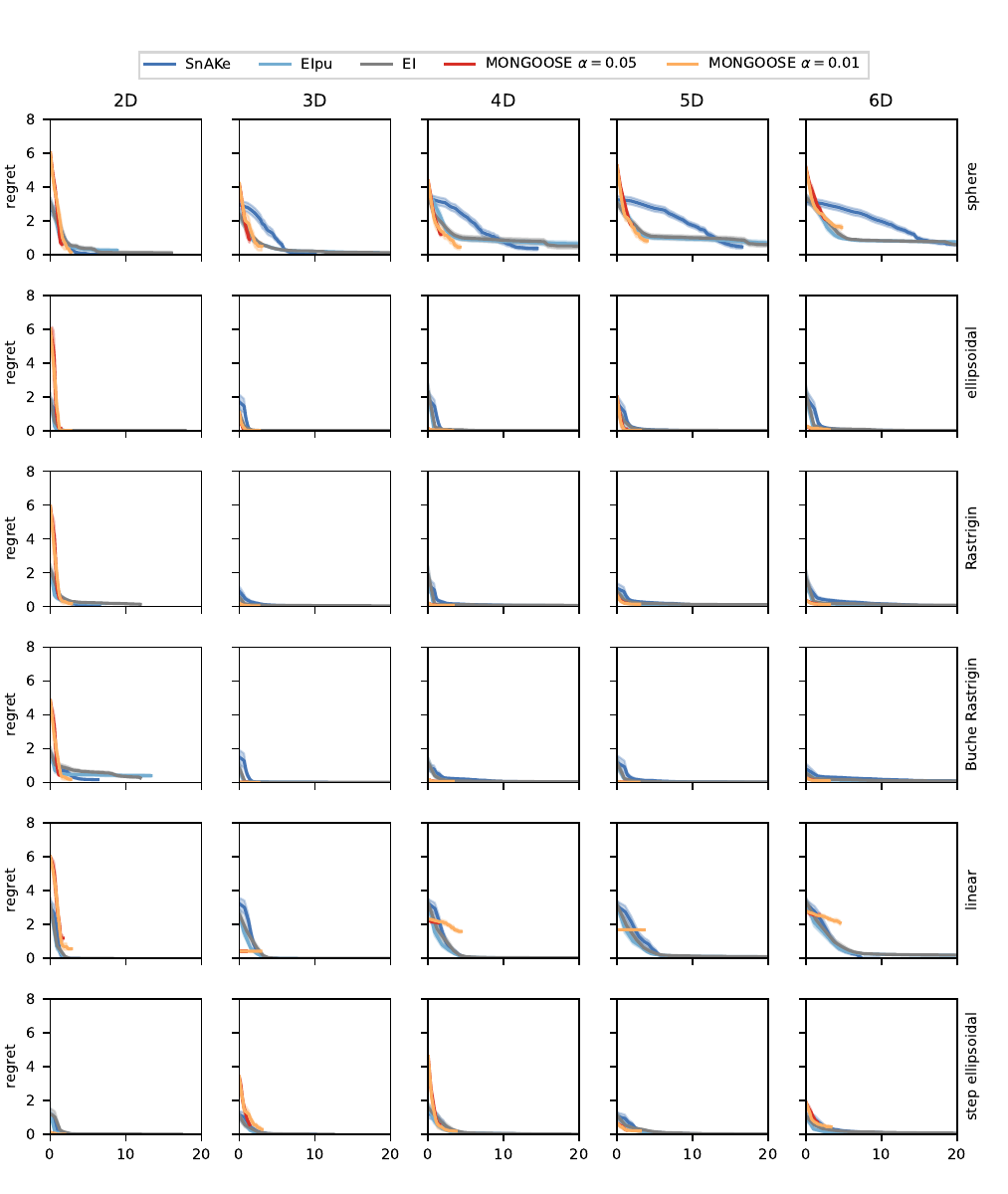}
    \caption{Individual COCO plots for Figure~\ref{fig:coco_avg}. COCO functions 1-6: sphere function, ellipsoidal function, Rastrigin function, B\"{u}che-Rastrigin function, linear slope, step ellipsoidal function.}
    \label{fig:coco_50_1}
\end{figure}

\begin{figure}
    \centering
    \includegraphics{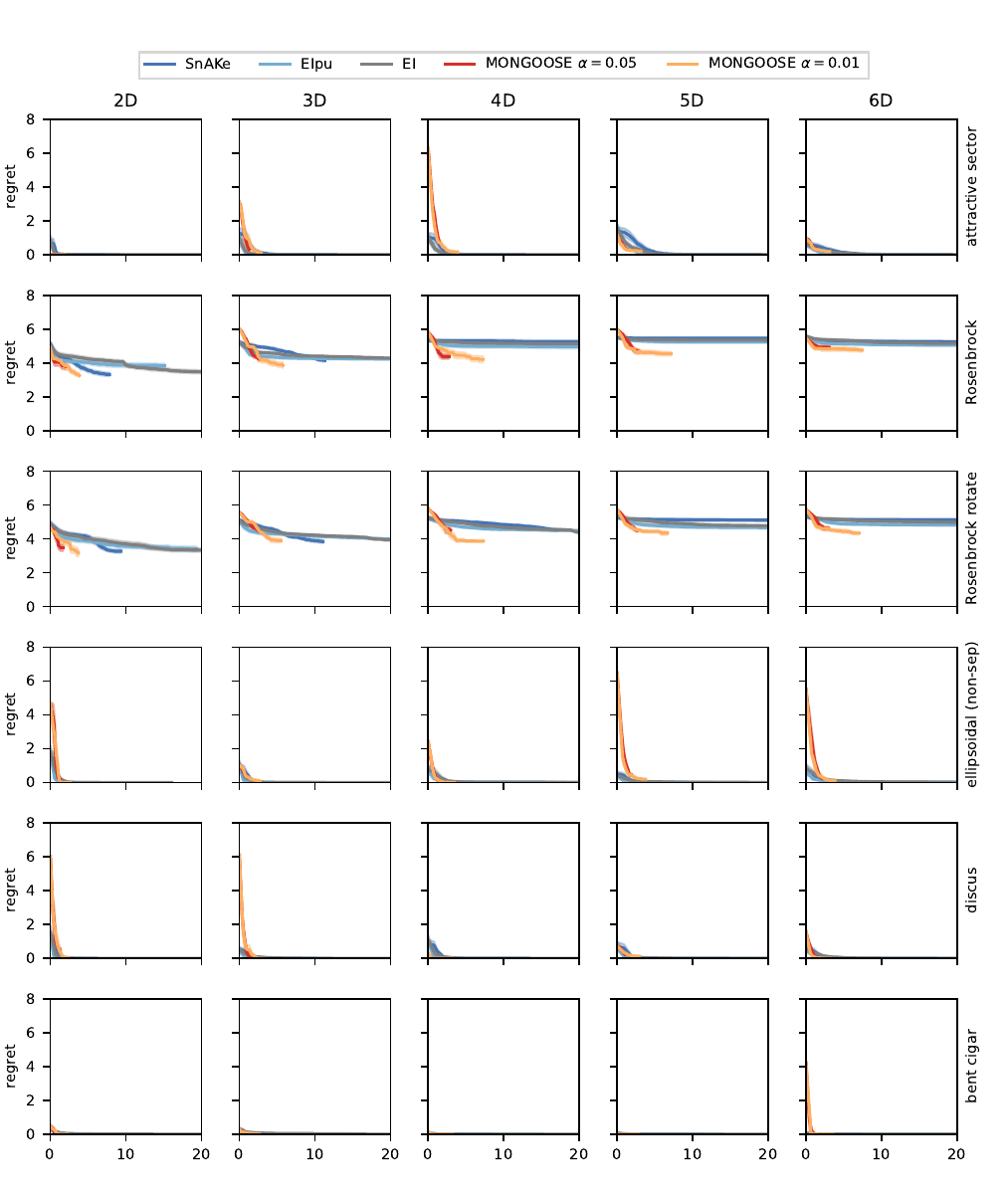}
    \caption{Individual COCO plots for Figure~\ref{fig:coco_avg}. COCO functions 7-12: attractive sector function, Rosenbrock (original) function, Rosenbrock (rotated) function, ellipsoidal (non-separable) function, discus function, bent cigar function.}
    \label{fig:coco_50_2}
\end{figure}

\begin{figure}
    \centering
    \includegraphics{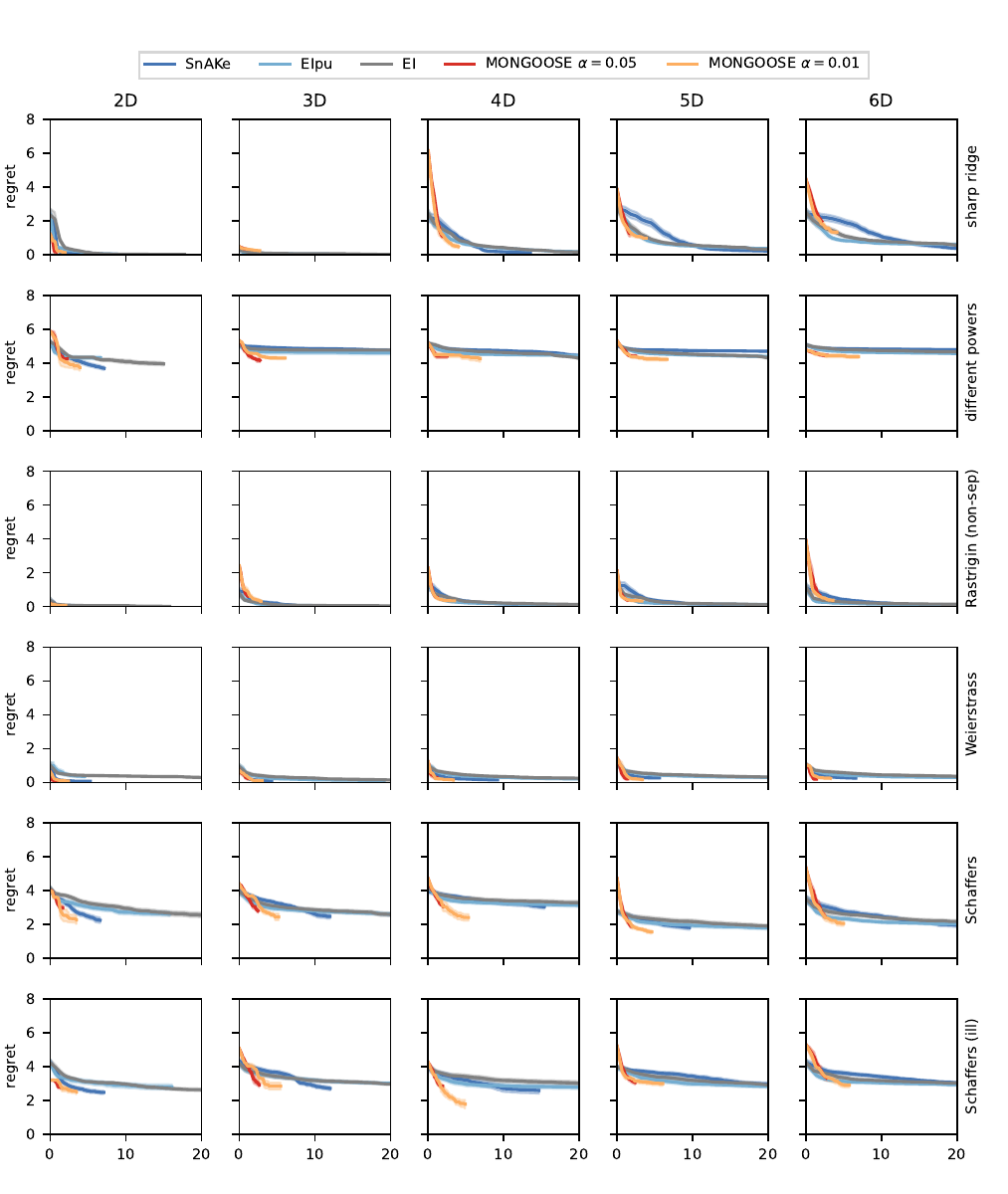}
    \caption{Individual COCO plots for Figure~\ref{fig:coco_avg}. COCO functions 13-18: sharp ridge function, different powers function, Rastrigin (non-separable) function, Weierstrrass function, Schaffers F7 function, Schaffers F7 (moderately ill-conditioned) function.}
    \label{fig:coco_50_3}
\end{figure}

\begin{figure}
    \centering
    \includegraphics{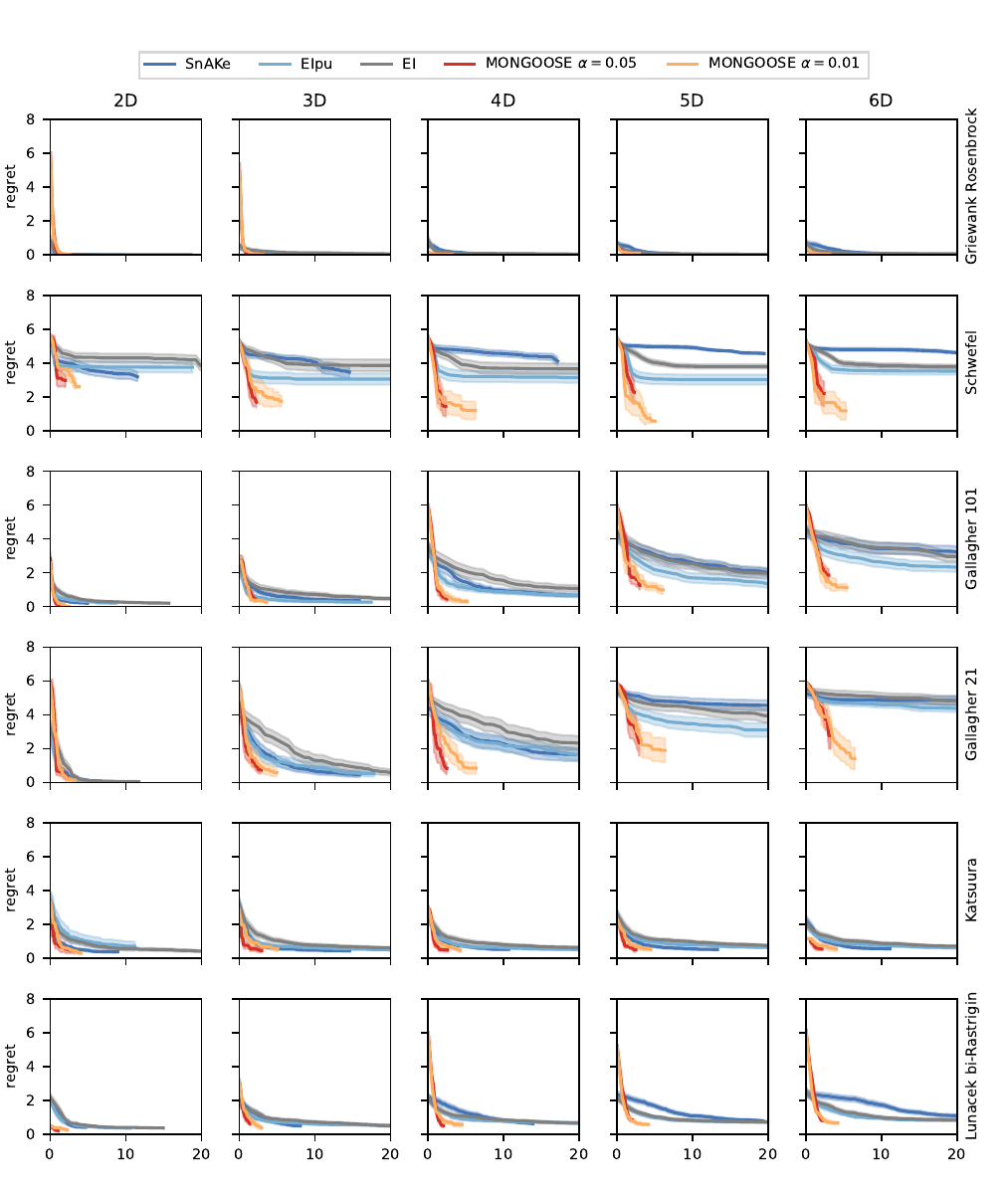}
    \caption{Individual COCO plots for Figure~\ref{fig:coco_avg}. COCO functions 19-24: composite Griewank-Rosenbrock function, Schwefel function, Gallagher's Gaussian 101-me peaks function, Gallagher's Gaussian 21-hi peaks function, Kastsuura function, Lunacek bi-Rastrigin function.}
    \label{fig:coco_50_4}
\end{figure}

\end{document}